\pdfoutput=1
\documentclass[11pt]{article}

\usepackage{acl}

\usepackage{times}
\usepackage{latexsym}
\usepackage{comment} 
\usepackage[T1]{fontenc}
\usepackage[utf8]{inputenc}     % allow utf-8 input
\usepackage[T1]{fontenc}        % use 8-bit T1 fonts
\usepackage{hyperref}           % hyperlinks
\usepackage{url}                % simple URL typesetting
\usepackage{booktabs}           % professional-quality tables
\usepackage{amsfonts}           % blackboard math symbols
\usepackage{amsmath}            % equation splits
\usepackage{nicefrac}           % compact symbols for 1/2, etc.
\usepackage{microtype}          % microtypography
\usepackage{graphicx}           % fancy tables
\usepackage{array}              % fancy tables part 2
\usepackage{longtable}          % fancy tables part 3
\usepackage{multirow}           % fancy tables part 4
\usepackage{caption}            % subfigures
\usepackage{subcaption}         % subfigures
\usepackage{amssymb}            % checkmarks
\usepackage{bbm}                % fancy indicator function

\usepackage{tcolorbox}          % for conclusion boxes

\newcommand{\ifcomments}{\iftrue}

\newcommand{\tool}{\texttt{Sybil}}
\newcolumntype{C}[1]{>{\centering\arraybackslash}p{#1}}
\newcommand*{\affaddr}[1]{#1} 
\newcommand*{\affmark}[1][*]{\textsuperscript{#1}}
\newcommand*{\email}[1]{\texttt{#1}}

\usepackage{microtype}

\title{Sibylvariant Transformations for Robust Text Classification}

\author{%
Fabrice Harel-Canada\affmark[1], Muhammad Ali Gulzar\affmark[2], Nanyun Peng\affmark[1], Miryung Kim\affmark[1] \\
\affaddr{\affmark[1]Computer Science Department, University of California, Los Angeles} \\
\affaddr{\affmark[2]Computer Science Department, Virginia Tech} \\
\email{\{fabricehc, violetpeng, miryung\}@cs.ucla.edu}, \email{gulzar@cs.vt.edu}
}

\begin{document}
\maketitle

\begin{abstract}
    The vast majority of text transformation techniques in NLP are inherently limited in their ability to expand input space coverage due to an implicit constraint to preserve the original class label. In this work, we propose the notion of \emph{sibylvariance (SIB)} to describe the broader set of transforms that relax the label-preserving constraint, knowably vary the expected class, and lead to significantly more diverse input distributions. We  offer a unified framework to organize all data transformations, including two types of SIB: (1) {\em Transmutations} convert one discrete kind into another, (2) {\em Mixture Mutations} blend two or more classes together. To explore the role of sibylvariance within NLP, we implemented 41 text transformations, including several novel techniques like \texttt{Concept2Sentence} and \texttt{SentMix}. Sibylvariance also enables a unique form of adaptive training that generates new input mixtures for the most confused class pairs, challenging the learner to differentiate with greater nuance. Our experiments on six benchmark datasets strongly support the efficacy of sibylvariance for generalization performance, defect detection, and adversarial robustness.
\end{abstract}

\section{Introduction}
\label{introduction}

Automatically generating new data is a critical component within modern machine learning pipelines. During training, data augmentation can expose models to a larger portion of  potential input space, consistently leading to better generalization and performance \cite{trans_inv, alexnet, effectiveness_da}. After training, creating effective test instances from existing data can expose specific model failure modes and provide actionable corrective feedback \cite{ml_test_survey, checklist}. 

While many techniques can artificially expand labeled training sets or test suites, nearly all of them are class-preserving. That is to say, the model outputs are invariant (INV) with respect to the transformations. This cautious constraint ensures the new data does not lie in an out-of-distribution null class which might impede the learning objective. However, it also requires more conservative transforms that inherently limit the degree of diversification. 

In this work, we propose and extensively investigate the potential of {\em sibylvariant (SIB)} transformations that {\em knowably} vary the expected class. From the Greek \emph{sibyls}, or oracles, the term parallels the oracle construction problem in software testing \cite{oracle_prob}. In a nutshell, sibylvariants either fully transmute a datum from one class $c_i$ to another $c_j$, or mix data from multiple classes together to derive a new input with a soft label that reflects the mixed membership. In this way, SIB can more strongly perturb and diversify the underlying distribution. Moreover, SIB makes possible a new type of adaptive training by synthesizing data from frequently confused class pairs, challenging the model to differentiate with greater refinement. 

In the following sections, we position SIB within a broader conceptual framework for all data transforms (Section \ref{sibylvariance}) and highlight several newly proposed techniques (Section \ref{transformations}). To support a comprehensive evaluation of how SIB may complement or even surpass its INV counterparts, we implemented 41 new and existing techniques into an open source tool called \texttt{Sibyl}. Equipped with the framework and tool, we evaluate 3 central research questions:
\begin{itemize}
    \itemsep0em 
    % \vspace{-0.70em}
    \item\textbf{RQ1. Generalization Performance.} Does training on SIB-augmented data improve model accuracy on the original test set? 
    %\vspace{-0.75em}
    \item\textbf{RQ2. Defect Detection.} For models trained on the original dataset, how effective are SIB tests at inducing misclassifications? 
    %\vspace{-0.75em} 
    \item\textbf{RQ3. Adversarial Robustness.} Are models trained on SIB-augmented data more robust to existing adversarial attack algorithms? \vspace{-0.5em}
\end{itemize} 
Our comprehensive evaluation encompasses 6 text classification datasets, 11 transformation pipelines, and 3 different levels of data availability. In total, we trained 216 models and generated over 30 million new training inputs, 480,000 testing inputs, and 3,300 adversarial inputs. In the generalization study, SIB attained the highest accuracies in 89\% (16 out of 18) of experimental configurations, with the adaptive mixture mutations being the most consistently effective. SIB also revealed the greatest number of model defects in 83\% (5 out of 6) of the testing configurations. Lastly, of all the experimental configurations where adversarial robustness was improved over the no-transform baseline, 92\% (11 out of 12) of them involved SIB. Overall, our findings strongly support the efficacy of sibylvariance for generalization performance, defect detection, and adversarial robustness. 

Lastly, we describe how SIB may operate theoretically and discuss potential threats to validity (Section \ref{discussion}) before contrasting it with related work (Section \ref{related_work}). The source code for \texttt{Sibyl} and our experiments is available at: \url{https://github.com/UCLA-SEAL/Sibyl}.

% \begin{center}
%     \url{https://zenodo.org/record/4835770} \fhnote{Replace this!}
% \end{center}
\section{Sibylvariance}
\label{sibylvariance}

% Words serve as guideposts to the ideas we wish to revisit and we'll need to define a few to navigate the remainder of the paper. Due the prevalence of classification tasks in ML, much of the language defaults to talk of \emph{classes} and \emph{labels}. \mgnote{@Fabrice: Should we just stick to class and labels?} In order to be inclusive of other task settings, the general language we use for these concepts are \emph{kinds} and \emph{behaviors}, respectively. 

All data transformations in the classification setting can be categorized into one of two types:

\begin{itemize}
    \itemsep0em 
    \item \textbf{\textit{Invariant (INV)}} preserves existing labels. 
    \vspace{-0.5em}
    \begin{equation} \label{eq1}
        \begin{split}
            \{T_{INV}(X_i), y_i\} & \rightarrow \{X_j, y_i\}  \\
            \text{where } X_i & \neq X_j
        \end{split}
        \vspace{-0.75em}
    \end{equation}
    For example, contracting ``What is the matter?'' to ``What's the matter?''should preserve a model behavior for sentiment analysis.
    
    \item \textbf{\textit{Sibylvariant (SIB)}} changes an existing label in a knowable manner.  
    \vspace{-0.5em}
    \begin{equation} \label{eq2}
        \begin{split}
            T_{SIB}(\{X_i, y_i\}) & \rightarrow \{X_j, y_j\} \\
            \text{where } X_i \neq X_j & \text{ and } y_i \neq y_j.
        \end{split}
    \end{equation}
    \vspace{-1.5em} 
    
    SIB transforms both the input $X_i$ to $X_j$ and the output label from $y_i$ to $y_j$ label, corresponding to the new $X_j$; such transformation is analogous to mutating an input and setting a corresponding oracle in metamorphic testing~\cite{metamorphic_testing}. For example, performing a verb-targeted antonym substitution on ``I love pizza.'' to generate ``I hate pizza.'' has the effect of negating the original semantics and will knowably affect the outcome of binary sentiment analysis.
\end{itemize}

It is important to note that transformation functions are not inherently INV nor SIB. The same exact transformation may have a different effect on expected model behavior depending on the particular classification task. For example, random word insertions generally have an INV effect on topic classification tasks, but would be SIB with respect to grammaticality tasks \cite{cola}. 

% Each transformation serves the role of either INV or SIB depending on a given classification task.  For example, constructing a mirrored image of the original driving scene would have an invariant effect for an object detection task, but would result in an opposite steering angle prediction, because a left-hand turn should be converted to a right-hand turn. %\mknote{having a vision example is fine, but I suggest another example relevant to text classification here.}

\subsection{Sibylvariant Subtypes}
SIB can be further refined based on the types and degree of semantic shift in newly generated data: 

\begin{itemize}
    \itemsep0em 
    \item \textbf{\textit{Transmutation}} changes one discrete kind into another, excluding the existing label, $L \backslash \{y_i\}$,
    \vspace{-1.5em}
    \begin{equation} \label{eq3}
        \begin{split}
            T_{SIB}(\{X_i, y_i\}) & \rightarrow \{X_j, y_j\} \\
            \text{where } X_i \neq X_j & \text{ and } y_j \in L \backslash \{y_i\}.
        \end{split}
    \end{equation} 
    Critically, the newly created data points retain stylistic and structural elements of the original that help boost diversity.
    
    \item \textbf{\textit{Mixture Mutation}} mixes inputs from multiple classes and interpolates the expected behavior into a mixed label distribution (i.e. {\em soft label}). Equivalently, we have:
    
    \vspace{-1.5em}
    \begin{equation}
        \begin{split}
            T_{SIB}(\{X_i, y_i\}) & \rightarrow \{X_j, y_j\} \\
            \text{where } X_i \neq X_j & \text{ and } y_j \in \bigcap_{l}^{|L|} \lambda_l
        \end{split}
    \end{equation} 
    \vspace{-1.5em}
    
    where the final term indicates a $\lambda$-degree of membership in each label $l$ belonging to the expected input space and is normalized as a probability distribution (i.e. $\sum_l \lambda_l = 1$). For example, a document with topic `surfing' can be combined with another document with topic `machine learning' to yield a new label with probability mass placed on both topics. While mixture mutations may seem unnatural, the intuition is that humans can recognize mixed examples and adjust their predictions accordingly. Models ought to do the same. 

    % \item \textbf{\textit{Directional Expectation}}\footnote{This kind of transformation was recently implemented in CheckList \cite{checklist}, which won best paper at ACL 2020.}: shifts the \emph{confidence} in the original behavior towards a different expectation. For example, within sentiment analysis, adding the phrase ``That being said, I can't say I'm happy about it.'' to any positive sentence should decrease the confidence in a prediction of ``positive'' even if that transformation does not succeed in fully transmuting the label to negative. 
\end{itemize}

\subsection{Adaptive Sibylvariant Training} 
\label{ada}
One unique and promising aspect of SIB is to target specific class pairings dynamically during training. In much the same way that a human teacher might periodically assess a students' understanding and alter their lesson plan accordingly, {\tool} computes a confusion matrix and constructs more examples containing classes for which the model has the most difficulty differentiating. For example, if a topic model most frequently misclassifies `science' articles as `business,' \emph{adaptive} SIB (denoted as $\alpha$SIB) will generate new blended examples of those classes in every mini-batch until the next evaluation cycle. At that point, if the model confuses `science' for ``health,''  $\alpha$SIB will construct new mixtures of those classes and so on. {\tool} supports built-in runtime monitoring for $\alpha$SIB training.

\section{Transformations}
\label{transformations}

In {\tool}, we defined 18 new transforms and adapt 23 existing techniques from prior work \cite{checklist, textattack, eda} to expand the coverage of SIB and INV text transformations. At a high level, Table \ref{fig:sibyl_transforms} shows these 41 transforms organized into 8 categories: {\tt Mixture} (i.e., blending text), {\tt Generative} (i.e. concept-based text generation), {\tt Swap} (e.g., substituting antonyms, synonyms, hypernyms, etc.), {\tt Negation} (e.g., adding or removing negation), {\tt Punctuation} (e.g., adding or removing punctuation), {\tt Text Insert} (e.g., adding negative, neutral, or positive phrases), {\tt Typos} (e.g. adding various typos), and {\tt Emojis} (e.g. adding or removing positive or negative emoji). We highlight several signature transforms here and provide a more detailed listing in Appendix \ref{appendix: transformations}.  

%See Appendix \ref{appendix: transformations} for brief descriptions of other possible transformations and Appendix \ref{appendix: sib_subtypes}: Table \ref{tab:sib_subtypes} for additional examples.
% As a result of an informal survey of related data augmentation literature and from our own creations, we identified 72 possible transformations to make upon text data (see Appendix \ref{appendix: transformations} for a brief description of each). For our \texttt{Sibyl} tool, we have implemented only 36 of them due to time and resource constraints, but view the remainder as viable directions for future research. It is not be feasible to review all of them here, but we will highlight a handful of the more interesting ones.

% \textbf{\texttt{ImportLinkText.}} HTTP links are generally low-information strings and difficult for models to use and understand effectively. However, the web pages accessed by the link provide a much more detailed account of the meaning conveyed when a human shares a link. This transformation finds HTTP links via regex pattern matching and then uses \texttt{urllib} to extract the visible page text and import it into the original text string. For example, a text input which reads "My thoughts on the '90s Godzilla movie: \url{ https://www.rogerebert.com/reviews/godzilla-1998}" would be difficult to classify as positive or negative in sentiment. However, with \texttt{ImportLinkText}, it becomes abundantly clear that the sentiment is negative. 

\begin{table}[htbp]
    \centering
    \small
    \resizebox{\columnwidth}{!}{%
    \begin{tabular}{|l|p{5cm}|}
    \hline
    \rowcolor[HTML]{EFEFEF} 
    \textbf{Category} & \textbf{Transformations}                                                                                                                                    \\ \hline
    Mixture           & TextMix$\dagger$, SentMix$\dagger$, WordMix$\dagger$                                                                                                                                   \\ \hline
    Generative        & Concept2Sentence$\dagger$, ConceptMix$\dagger$       
                                                                                        \\ \hline
    
    Swap         & ChangeNumber, ChangeSynonym, ChangeAntonym, ChangeHyponym, ChangeHypernym, ChangeLocation, ChangeName, RandomSwap                                           \\ \hline
    Negation          & AddNegation, RemoveNegation                                                                                                                                 \\ \hline
    Punctuation       & ExpandContractions, ContractContractions                                                                                                                    \\ \hline
    Text Insert    & RandomInsertion, AddPositiveLink$\dagger$, AddNegativeLink$\dagger$, ImportLinkText$\dagger$, InsertPositivePhrase, InsertNegativePhrase                                               \\ \hline
    Typos             & RandomCharDel, RandomCharInsert, RandomCharSubst, RandomCharSwap, RandomSwapQwerty, WordDeletion, HomoglyphSwap                                             \\ \hline
    Emojis            & Emojify$\dagger$, AddEmoji$\dagger$, AddPositiveEmoji$\dagger$, AddNegativeEmoji$\dagger$, AddNeutralEmoji$\dagger$, Demojify$\dagger$, RemoveEmoji$\dagger$, RemovePositiveEmoji$\dagger$, RemoveNegativeEmoji$\dagger$, RemoveNeutralEmoji$\dagger$ \\ \hline
    \end{tabular}%
    }
    \caption{Transformations currently available in {\tool}. New transforms that we defined are marked with $\dagger$.}
    \label{fig:sibyl_transforms}
    \vspace{-2em}
\end{table}

\paragraph{Concept2Sentence (C2S).} C2S is a two step process: (1) extract a short list of key concepts from a document and (2) generate a new sentence that retains critical semantic content of the original while varying its surface form, style, and even subject matter. To accomplish this, we leveraged integrated gradients \cite{integrated_gradients, Pierse_Transformers_Interpret_2021} to produce saliency attributions that identify the most relevant tokens for a given class label. We then generate a well-composed sentence from the extracted concepts using a pre-trained BART \cite{BART} model fine-tuned on the CommonGen dataset \cite{CommonGen}.

\begin{figure}[t]
    \centering
    \includegraphics[width=0.475\textwidth]{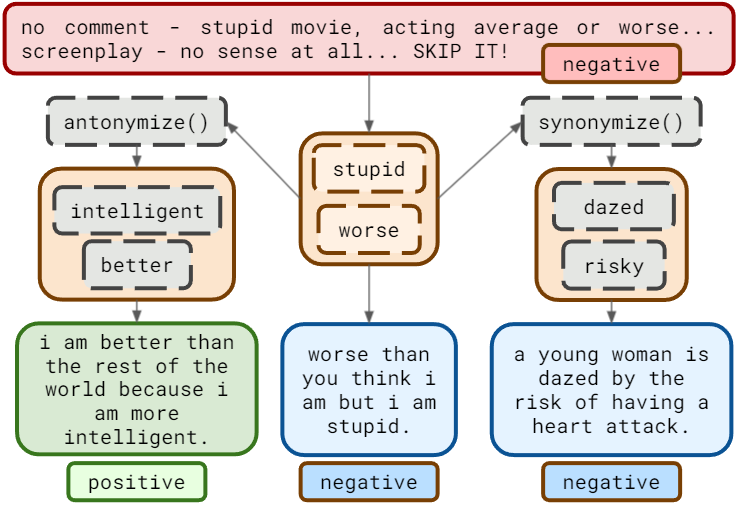}
    \vspace{-0.5ex}
    \caption{C2S intakes a text and its label (red) to extract keywords, ['stupid, worse']. These words are used to generate a new INV sentence shown in red. Alternatively, antonym (left) and synonym (right) substitution can produce new concepts that further boost diversity.}
    \label{fig:c2s}
    \vspace{-1em}
\end{figure}

Prior to generation, it is possible to apply other transformations to the extracted concepts to encourage diversity or knowably alter the label. For example, on the left hand side of Figure \ref{fig:c2s} an antonym substitution produces a SIB effect by changing the extracted concepts from ['stupid', 'worse'] to ['intelligent', 'better']. The new sentence exhibits a change in subject and style, but is correctly transmuted to have positive sentiment. C2S is thus an extremely promising transformation for diversifying text along both INV and SIB directions.  %MK: I like the new example on stupid and intelligent to dazed and risky, etc. Good! FHC: Thanks! :)

\paragraph{TextMix, SentMix, and WordMix.} Mixture mutations, like mixup \cite{mixup} and cutmix \cite{cutmix} from the image domain, take a batch of inputs and blend them together to form new inputs with an interpolated loss and they have shown robustness to adversarial attacks. \texttt{TextMix} translates this idea to the text domain by merging two inputs and interpolating a \emph{soft label} according to the proportion of tokens belonging to the constituent classes. While \texttt{TextMix} does a straightforward concatenation, \texttt{SentMix} shuffles the sentences and thus encourages long-range comprehension. \texttt{WordMix} concatenates and shuffles all words, encouraging keyword-to-topic understanding when sentence structure is compromised.

\section{Experiments}
\label{experiments}

% OLD INTRO
% In order to determine the extent to which sibylvariance can create effective tests which detect model defects and generate new training data that improves model performance, we investigate our three research questions. We empirically validate the proposed framework by \emph{a)} exploring natural language processing tasks and \emph{b)} focusing on the performance of classification models trained on labeled datasets augmented with our functions and others used in practice.

% We experimentally study our unified framework for data transformations

%\subsection{Datasets.} 
 % \mknote{FHC, How is 144 calculated?} \fhnote{144 = 6 datasets x 3 different numbers of classes x 8 different TPs. NOTE: We study 11 TPs in total, 6 x 3 x 11 = 198 different experimental configurations. However, 3 of them are adaptive pipelines and we don't create / store that data - it is generated at run-time and then discarded.}

\subsection{Transformation Pipelines \& Datasets} 

To compare the potential of INV, SIB, and both (INVSIB) in aggregate, we construct a {\em transformation pipeline}  ($TP$)~\cite{randaugment, uda}, where we uniformly sample $n$ transformations of the selected kind to generate new $\{X_i,y_i\}$ pairs. We also create $TP$s that apply a single transform, $T_{\texttt{SINGLE}}$, to highlight the efficacy of {\tt C2S}, {\tt TextMix}, {\tt SentMix}, {\tt WordMix} and their adaptive versions, prefixed with $\alpha$. In total, we evaluate 11 $TP$s per dataset, shown in Table \ref{tab:trans_pipelines}. 

Due to space limitations, we report the top performing $TP$ of each kind using an asterisk (*). INV* represents the best from $T_{\text{INV}}$ and $T_{\text{C2S}}$, while SIB* represents the best from $T_{\text{SIB}}$ and the mixture mutations. For \textbf{RQ1}, we also compare against TMix \cite{mixtext}, EDA \cite{eda}, and AEDA \cite{aeda}. TMix is a recent \emph{hidden}-space mixture mutation for text, as opposed to {\tool}'s direct mixture mutation on the input space with greater transparency and examinability. EDA and AEDA are examples of recent INV transformations. Full results are available in the appendices.% MK: I wanted to see deeper discusson on TMix, but since you don't have it. I am going to keep what I had earlier 

\begin{table}[htbp]
    \centering
    \vspace{-0.35em}
    \begin{tabular}{l p{5cm}}
        \toprule
        \textbf{Shorthand} & \textbf{Description}             \\
        \midrule
        $T_{\text{ORIG}}$      & 0 transformations as baseline     \\
        $T_{\text{INV}}$       & sample 2 INVs             \\
        $T_{\text{SIB}}$       & sample 2 SIBs             \\
        $T_{\text{INVSIB}}$    & sample 1 INV and 1 SIB    \\
        $T_{\text{SINGLE}}$  & apply one from C2S, TextMix, SentMix, WordMix, $\alpha$TextMix, $\alpha$SentMix, $\alpha$WordMix \\
        \midrule
    \end{tabular}
    \vspace{-0.5em}
    \caption{$TP$ descriptions. $TP$s with an $\alpha$-prefix use targeted, adaptive training (Section \ref{ada}).}
    \label{tab:trans_pipelines}
    \vspace{-1em}
\end{table}

We study six benchmarks for two kinds of NLP tasks: topic classification and sentiment analysis. Table \ref{tab:datasets} summarizes their relevant details. To simulate different levels of resource availability, we create three data subsets with by varying number of examples per class --- 10, 200, and 2500. These subsets were expanded $30\times$ via augmentation for each $TP$. In total, we generated 144 new datasets (144 = 6 benchmarks * 3 levels of data availability * 8 $TP$s which persist data. $\alpha$SIB is runtime only.)

\begin{table*}[htbp]
    \centering
    \resizebox{\textwidth}{!}{%
    \begin{tabular}{l l l l c r r}
        \toprule
        \textbf{Dataset} & \textbf{Source}         & \textbf{Task}    & \textbf{Subject}      & \textbf{Classes}  & \textbf{Test} & \textbf{Avg Len} \\
        \midrule
        AG News          & \cite{CCN_datasets}     & Topic              & News Articles         & 4                 &  1,900    & 38  \\
        DBpedia          & \cite{CCN_datasets}     & Topic              & Wikipedia Articles    & 14                &  5,000    & 46  \\
        Yahoo! Answers   & \cite{CCN_datasets}     & Topic              & QA Posts              & 10                &  6,000    & 92  \\
        Amazon Polarity  & \cite{CCN_datasets}     & Sentiment          & Product Reviews       & 2                 &  200,000  & 74  \\
        Yelp Polarity    & \cite{CCN_datasets}     & Sentiment          & Business Reviews      & 2                 &  10,000   & 133 \\
        IMDB             & \cite{IMDB}             & Sentiment          & Movies Reviews        & 2                 &  12,500   & 234 \\
        \midrule
    \end{tabular}
    }
    \vspace{-1em}
    \caption{Dataset details. Test represents the number of examples per class in the test set. } 
    \label{tab:datasets}
    \vspace{-1em}
\end{table*}

\subsection{Model Setting}

We used a \texttt{bert-base-uncased} model~\cite{BERT} with average pooling of encoder output, followed by a dropout layer \cite{dropout} with probability $0.1$, and a single linear layer with hidden size 768 and GELU \cite{gelu} activation. Maximum sentence length was set to 250. We use a batch size $16$, an Adam optimizer \cite{adam} with a linear warmup, a $0.1$ weight decay, and compute accuracy every $2,000$ steps. All models were trained for $30$ epochs on eight Nvidia RTX A6000 GPUs, with early stopping. In total, we constructed 198 different models. 

For all $TP$s that produce a soft-label, we use a multi-class cross-entropy loss and  computed performance via a weighted top-k accuracy,
\vspace{-0.75em}
\begin{equation} \label{eq:top_k_acc}
    \sum_j^k \lambda_l \cdot \mathbbm{1}(y_l = \hat{y}_j),
    \vspace{-0.75em}
\end{equation}
where $\lambda_j$ is the degree of class membership, $\mathbbm{1}(\cdot)$ is the indicator function, and $y_j$ and $\hat{y}_j$ are the indices of the $j$-th largest predicted score for the ground truth label and predicted label, respectively. 

%For example, if a model predicted $[0.5, 0.1, 0.4]$ for the label $[0.7, 0.3, 0.0]$ and $k=2$, the weighted top-k accuracy is $0.7$ because the first index correctly overlaps, but the others do not.

\subsection{RQ1. Generalization Performance}
\label{RQ1}

For RQ1, we explore how model accuracy on the original test set is influenced by training data augmented with INV and SIB transformations. Table \ref{tab:train_acc} shows the results on six benchmarks with three levels of data availability. 

\begin{figure}[t!]
    \centering
    \includegraphics[width=\columnwidth]{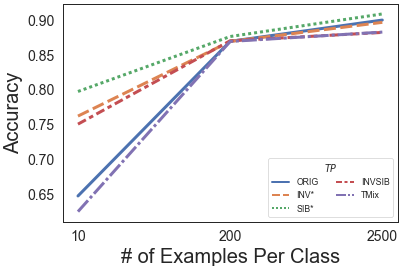}
    \vspace{-1.5em}
    \caption{SIB* outperforms INV* most, when data availability is low, indicating the necessity of SIB to complement INV.}
    \label{fig:accuracy_by_num_examples}
    \vspace{-0.5em}
\end{figure}

\begin{figure}[t!]
    \centering
    \includegraphics[width=\columnwidth]{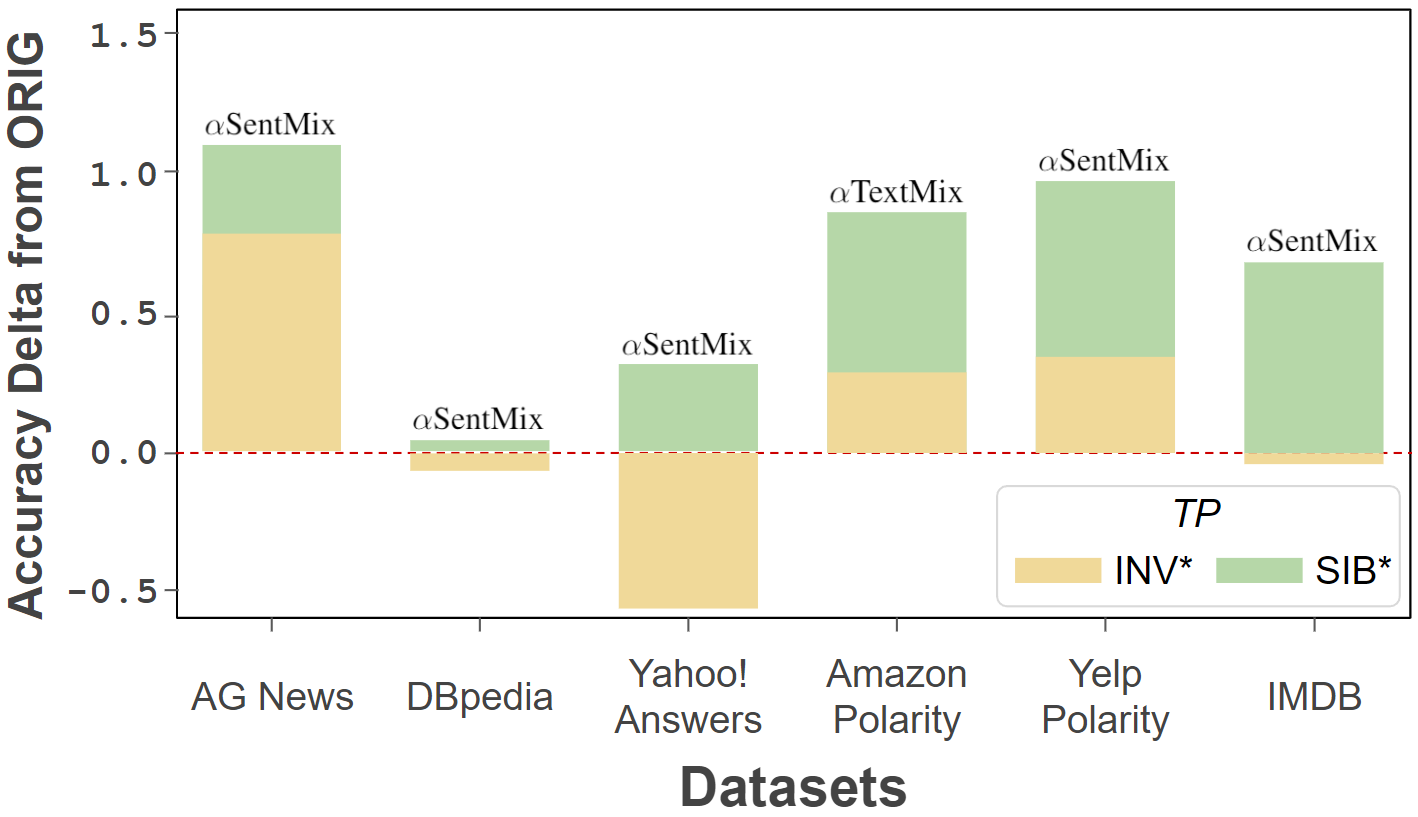}
    \vspace{-1.75em}
    \caption{The best performing TP for each dataset trained on 200 examples per class. $\alpha$SentMix or $\alpha$TextMix leads to the highest performance gains. SIB* consistently outperforms INV*.}
    \label{fig:accuracy_INV_vs_SIB}
    \vspace{-1.5em}
\end{figure}

We observe the most significant performance gains when training 10 examples per class \textemdash accuracy is improved by 4.7\% on average across all datasets and by a maximum of up to 15\% for IMDB. Figure \ref{fig:accuracy_by_num_examples} shows that as the number of labeled training data increases, a dominant trend emerged \textemdash $T_{\text{SIB}}$ always generalized better to unseen test data. In fact, the only kind of transformation to always outperform both $T_{\text{ORG}}$ and TMix is SIB*. Figure \ref{fig:accuracy_INV_vs_SIB} shows the performance delta between INV* and SIB* against the $T_{\text{ORG}}$ baseline at 200 examples per class. For every dataset, either $\alpha$SentMix or $\alpha$TextMix is the best performing $TP$, while INV* actually leads to performance decreases for DBPedia, Yahoo! Answers, and IMDB.

%\vspace{-0.5em}

%\vspace{-0.5em}

One key reason that aided SIB in attaining strong performance is the use of adaptive training. On average, crafting new examples that target the model's primary confusions during training added approximately 1\% to accuracy relative to mixing classes uniformly at random. This shows another unique benefit of sibylvariance that is not transferable to its INV counterparts. 

\begin{table*}[t]
\vspace{-1em}
\centering
\resizebox{\textwidth}{!}{%
\begin{tabular}{|c|c|ccc|c|c|ccc|c|c|ccc|}
\hline
\rowcolor[HTML]{EFEFEF} 
\cellcolor[HTML]{EFEFEF} & \cellcolor[HTML]{EFEFEF} & \multicolumn{3}{c|}{\cellcolor[HTML]{EFEFEF}\textbf{\# Examples / Class}} & \cellcolor[HTML]{EFEFEF} & \cellcolor[HTML]{EFEFEF} & \multicolumn{3}{c|}{\cellcolor[HTML]{EFEFEF}\textbf{\# Examples / Class}} & \cellcolor[HTML]{EFEFEF} & \cellcolor[HTML]{EFEFEF} & \multicolumn{3}{c|}{\cellcolor[HTML]{EFEFEF}\textbf{\# Examples / Class}} \\ \cline{3-5} \cline{8-10} \cline{13-15} 
\rowcolor[HTML]{EFEFEF} 
\multirow{-2}{*}{\cellcolor[HTML]{EFEFEF}\textbf{Dataset}} & \multirow{-2}{*}{\cellcolor[HTML]{EFEFEF}\textbf{TP}} & \multicolumn{1}{c|}{\cellcolor[HTML]{EFEFEF}\textbf{10}} & \multicolumn{1}{c|}{\cellcolor[HTML]{EFEFEF}\textbf{200}} & \textbf{2500} & \multirow{-2}{*}{\cellcolor[HTML]{EFEFEF}\textbf{Dataset}} & \multirow{-2}{*}{\cellcolor[HTML]{EFEFEF}\textbf{TP}} & \multicolumn{1}{c|}{\cellcolor[HTML]{EFEFEF}\textbf{10}} & \multicolumn{1}{c|}{\cellcolor[HTML]{EFEFEF}\textbf{200}} & \textbf{2500} & \multirow{-2}{*}{\cellcolor[HTML]{EFEFEF}\textbf{Dataset}} & \multirow{-2}{*}{\cellcolor[HTML]{EFEFEF}\textbf{TP}} & \multicolumn{1}{c|}{\cellcolor[HTML]{EFEFEF}\textbf{10}} & \multicolumn{1}{c|}{\cellcolor[HTML]{EFEFEF}\textbf{200}} & \textbf{2500} \\ \hline
 & ORIG & \multicolumn{1}{c|}{75.08} & \multicolumn{1}{c|}{88.70} & 91.65 &  & ORIG & \multicolumn{1}{c|}{95.71} & \multicolumn{1}{c|}{98.87} & 98.96 &  & ORIG & \multicolumn{1}{c|}{56.24} & \multicolumn{1}{c|}{69.77} & 73.18 \\ \cline{2-5} \cline{7-10} \cline{12-15} 
 & INV* & \multicolumn{1}{c|}{\textbf{84.28}} & \multicolumn{1}{c|}{89.46} & 91.95 &  & INV* & \multicolumn{1}{c|}{97.29} & \multicolumn{1}{c|}{98.81} & 99.00 &  & INV* & \multicolumn{1}{c|}{61.39} & \multicolumn{1}{c|}{69.21} & 72.53 \\ \cline{2-5} \cline{7-10} \cline{12-15} 
 & SIB* & \multicolumn{1}{c|}{83.52} & \multicolumn{1}{c|}{\textbf{89.80}} & \textbf{92.42} &  & SIB* & \multicolumn{1}{c|}{\textbf{97.96}} & \multicolumn{1}{c|}{\textbf{98.90}} & \textbf{99.06} &  & SIB* & \multicolumn{1}{c|}{\textbf{62.47}} & \multicolumn{1}{c|}{\textbf{70.10}} & \textbf{73.37} \\ \cline{2-5} \cline{7-10} \cline{12-15} 
 & INVSIB & \multicolumn{1}{c|}{84.09} & \multicolumn{1}{c|}{89.00} & 91.36 &  & INVSIB & \multicolumn{1}{c|}{95.64} & \multicolumn{1}{c|}{98.74} & 98.92 &  & INVSIB & \multicolumn{1}{c|}{62.01} & \multicolumn{1}{c|}{67.75} & 73.16 \\ \cline{2-5} \cline{7-10} \cline{12-15} 
 & TMix $\ddagger$ & \multicolumn{1}{c|}{81.38} & \multicolumn{1}{c|}{88.62} & 89.43 &  & TMix $\ddagger$ & \multicolumn{1}{r|}{97.51} & \multicolumn{1}{r|}{98.66} & \multicolumn{1}{r|}{98.89} &  & TMix $\ddagger$ & \multicolumn{1}{c|}{53.68} & \multicolumn{1}{c|}{69.03} & 69.50 \\ \cline{2-5} \cline{7-10} \cline{12-15} 
 & EDA $\ddagger$ & \multicolumn{1}{r|}{81.50} & \multicolumn{1}{r|}{88.98} & \multicolumn{1}{r|}{90.93} &  & EDA  $\ddagger$ & \multicolumn{1}{r|}{97.42} & \multicolumn{1}{r|}{98.63} & \multicolumn{1}{r|}{98.89} &  & EDA $\ddagger$ & \multicolumn{1}{r|}{57.88} & \multicolumn{1}{r|}{68.03} & \multicolumn{1}{r|}{69.15} \\ \cline{2-5} \cline{7-10} \cline{12-15} 
\multirow{-7}{*}{\textbf{AG News}} & AEDA $\ddagger$ & \multicolumn{1}{r|}{81.03} & \multicolumn{1}{r|}{88.74} & \multicolumn{1}{r|}{92.09} & \multirow{-7}{*}{\textbf{DBpedia}} & AEDA $\ddagger$ & \multicolumn{1}{r|}{97.30} & \multicolumn{1}{r|}{98.88} & \multicolumn{1}{r|}{98.89} & \multirow{-7}{*}{\textbf{\begin{tabular}[c]{@{}c@{}}Yahoo! \\ Answers\end{tabular}}} & AEDA $\ddagger$ & \multicolumn{1}{r|}{59.51} & \multicolumn{1}{r|}{67.37} & \multicolumn{1}{r|}{69.91} \\ \hline
 & ORIG & \multicolumn{1}{c|}{67.30} & \multicolumn{1}{c|}{89.22} & 92.08 &  & ORIG & \multicolumn{1}{c|}{74.62} & \multicolumn{1}{c|}{91.66} & 93.70 &  & ORIG & \multicolumn{1}{c|}{64.70} & \multicolumn{1}{c|}{86.96} & 90.02 \\ \cline{2-5} \cline{7-10} \cline{12-15} 
 & INV* & \multicolumn{1}{c|}{73.69} & \multicolumn{1}{c|}{89.53} & 92.21 &  & INV* & \multicolumn{1}{c|}{\textbf{83.91}} & \multicolumn{1}{c|}{92.00} & 94.29 &  & INV* & \multicolumn{1}{c|}{76.20} & \multicolumn{1}{c|}{86.94} & 89.69 \\ \cline{2-5} \cline{7-10} \cline{12-15} 
 & SIB* & \multicolumn{1}{c|}{\textbf{74.90}} & \multicolumn{1}{c|}{\textbf{90.03}} & \textbf{92.26} &  & SIB* & \multicolumn{1}{c|}{80.46} & \multicolumn{1}{c|}{\textbf{92.60}} & \textbf{94.69} &  & SIB* & \multicolumn{1}{c|}{\textbf{79.74}} & \multicolumn{1}{c|}{\textbf{87.65}} & \textbf{90.90} \\ \cline{2-5} \cline{7-10} \cline{12-15} 
 & INVSIB & \multicolumn{1}{c|}{73.50} & \multicolumn{1}{c|}{89.06} & 91.26 &  & INVSIB & \multicolumn{1}{c|}{78.90} & \multicolumn{1}{c|}{91.85} & 93.03 &  & INVSIB & \multicolumn{1}{c|}{75.04} & \multicolumn{1}{c|}{87.04} & 88.24 \\ \cline{2-5} \cline{7-10} \cline{12-15} 
 & TMix $\ddagger$ & \multicolumn{1}{c|}{62.14} & \multicolumn{1}{c|}{87.98} & 91.00 &  & TMix $\ddagger$ & \multicolumn{1}{c|}{61.81} & \multicolumn{1}{c|}{91.19} & 92.80 &  & TMix $\ddagger$ & \multicolumn{1}{c|}{62.45} & \multicolumn{1}{c|}{86.94} & 88.29 \\ \cline{2-5} \cline{7-10} \cline{12-15} 
 & EDA $\ddagger$ & \multicolumn{1}{r|}{59.40} & \multicolumn{1}{r|}{87.68} & \multicolumn{1}{r|}{92.20} &  & EDA $\ddagger$ & \multicolumn{1}{r|}{71.90} & \multicolumn{1}{r|}{90.88} & \multicolumn{1}{r|}{94.11} &  & EDA $\ddagger$ & \multicolumn{1}{r|}{67.37} & \multicolumn{1}{r|}{86.45} & \multicolumn{1}{r|}{89.07} \\ \cline{2-5} \cline{7-10} \cline{12-15} 
\multirow{-7}{*}{\textbf{\begin{tabular}[c]{@{}c@{}}Amazon\\ Polarity\end{tabular}}} & AEDA $\ddagger$ & \multicolumn{1}{r|}{64.72} & \multicolumn{1}{r|}{88.92} & \multicolumn{1}{r|}{91.83} & \multirow{-7}{*}{\textbf{\begin{tabular}[c]{@{}c@{}}Yelp\\ Polarity\end{tabular}}} & AEDA $\ddagger$ & \multicolumn{1}{r|}{79.39} & \multicolumn{1}{r|}{91.60} & \multicolumn{1}{r|}{94.06} & \multirow{-7}{*}{\textbf{IMDB}} & AEDA $\ddagger$ & \multicolumn{1}{r|}{72.61} & \multicolumn{1}{r|}{86.56} & \multicolumn{1}{r|}{88.63} \\ \hline
\end{tabular}%
}
\vspace{-0.5em}
\caption{RQ1 accuracy comparison for INV*, SIB*, and INVSIB against baselines ORIG, TMix~\cite{mixtext}, EDA \cite{eda}, AEDA \cite{aeda}. An asterisk (*) indicates the best performance observed across underlying $TP$s of each kind, while a $\ddagger$ indicates related works for comparison.}
\label{tab:train_acc}
\vspace{-1.25em}
\end{table*}

% \mknote{take a look my edit below}
While our full scale experiments show a clear trend that SIB generally outperforms INV, we primarily evaluated $TP$s combining multiple transforms instead of assessing the efficacy of each in isolation. Initially, this was a logistical decision due to computational limitations. To investigate each transformation's effect individually, we conducted a small scale experiment training 756 models (($39$ transformations + $3$ $\alpha$SIB) $\times$ $6$ datasets $\times$ $3$ runs) on 10 examples per class with a $3\times$ augmentation multiplier. Based on this experiment, we then computed each transform's performance by averaging the accuracy change relative to a $T_{\text{ORIG}}$ baseline across all datasets. Table \ref{tab:top10_trans} shows the top ten best performing transforms, six of which employ SIB techniques. These results expand support for the overall conclusion that sibylvariance represents an especially effective class of transformations for improving generalization performance. 

\begin{table}[htbp]
\centering
\resizebox{\columnwidth}{!}{%
\begin{tabular}{|l|l|c|}
\hline
\rowcolor[HTML]{EFEFEF}
\textbf{Transform}   & \textbf{Type} & \textbf{Avg $\Delta$ (\%)} \\ \hline
$\alpha$SentMix      & SIB           & +4.26              \\ \hline
$\alpha$TextMix      & SIB           & +3.55              \\ \hline
RandomCharInsert     & INV           & +3.55              \\ \hline
TextMix              & SIB           & +3.22              \\ \hline
Concept2Sentence     & INV           & +2.70              \\ \hline
AddPositiveLink      & INV / SIB     & +2.48              \\ \hline
AddNegativeEmoji     & INV / SIB     & +2.45              \\ \hline
SentMix              & SIB           & +2.33              \\ \hline
ExpandContractions   & INV           & +2.15              \\ \hline
RandomCharSubst      & INV           & +2.06              \\ \hline
\end{tabular}%
}
\vspace{-0.5em}
\caption{Top ten individual transforms over a no-transform baseline averaged across all datasets. The INV / SIB types were SIB for the sentiment analysis datasets and INV for the topic classification datasets. See Table \ref{tab:training_acc_by_transform} in the Appendix for more details.}
\label{tab:top10_trans}
\vspace{-1.6em}
\end{table}

% \begin{table}[htbp]
% \centering
% %\resizebox{\columnwidth}{!}{%
% \begin{tabular}{|l|l|r|}
% \hline
% \textbf{Transform}   & \textbf{Type} & \textbf{Rank} \\ \hline
% $\alpha$TextMix      & SIB           & 1             \\ \hline
% $\alpha$SentMix      & SIB           & 2             \\ \hline
% TextMix              & SIB           & 2             \\ \hline
% AddNegativeEmoji     & INV / SIB     & 4             \\ \hline
% RandomCharInsert     & INV           & 4             \\ \hline
% AddPositiveLink      & INV / SIB     & 7             \\ \hline
% AddNegativeLink      & INV / SIB     & 8             \\ \hline
% Concept2Sentence     & INV           & 9             \\ \hline
% SentMix              & SIB           & 10            \\ \hline
% InsertNegativePhrase & INV / SIB     & 10            \\ \hline
% \end{tabular}
% \vspace{-0.5em}
% \caption{Top ten transformations by rank of average accuracy increase across all datasets. All INV / SIB types were SIB for the sentiment analysis datasets and INV for the topic classification datasets.}
% \label{tab:top10_trans}
% \vspace{-1.25em}
% \end{table}

\vspace{0.5em}
\begin{tcolorbox}
    \textbf{Generalization Performance.} \\
    Models trained upon SIB-augmented data attained the highest test set accuracy in 89\% (16 out of 18) of experimental configurations, with the adaptive mixture mutations being the most consistently effective.
\end{tcolorbox}

\subsection{RQ2. Defect Detection}
\label{RQ2}

For RQ2, we assess how generating new tests with INV and SIB can expose defective model behavior. A single test is simply an $\{X_i,y_i\}$ pair and a test suite is a set of such tests. Defective behavior is misclassification, which is measured via a test suite's accuracy. For each dataset $D$, we select a high-performing BERT model trained only on the original dataset without any augmentation. Then for each of eight $TP$s (excluding $\alpha$SIB relevant to training only), we create 100 test suites, each containing 100 randomly sampled tests. This yields a total of 480,000 tests. We then report an average accuracy for each $D$ and $TP$ pair.

% toning down the testing claim wasn't as compelling to me as justifying it above... 

% \textcolor{blue}{Tests which lie outside expected input distribution are not likely to be fair nor actionable. Since the space of natural language is intractably large, determining exactly where the boundaries exist is challenging. We propose the following heuristic: if training on a set of $\{X_i,y_i\}$ pairs can improve model generalization performance on known, in-domain data, then the set can be considered a valid and useful test suite. Consequently, the positive results of the generalization study in \textbf{RQ1} supports the use of SIB transformations as reasonable for testing. After all, if sibylvariant transformations pushed the new data out of distribution, it is difficult to see how they would so consistently improve performance on in-domain examples.}

% FHC: I'm going to remove this paragraph for now, keeping it in case the ACL reviewers doubt the validity of the defect detection study like the NeurIPS reviewers did

\begin{figure}[t!]
    \centering
    \begin{subfigure}[b]{0.23\textwidth}
        \centering
        \includegraphics[width=\textwidth]{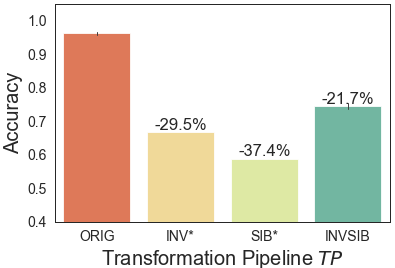}
        \caption{AG News}
        \label{fig:test_acc_ag_news}
    \end{subfigure}
    \begin{subfigure}[b]{0.23\textwidth}
        \centering
        \includegraphics[width=\textwidth]{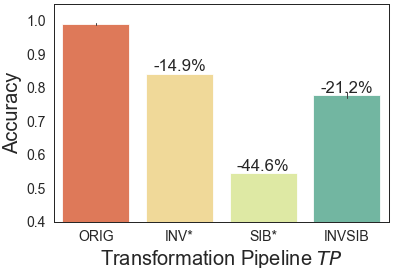}
        \caption{DBpedia}
        \label{fig:test_acc_dbpedia_14}
    \end{subfigure}
    \\
    \begin{subfigure}[b]{0.23\textwidth}
        \centering
        \includegraphics[width=\textwidth]{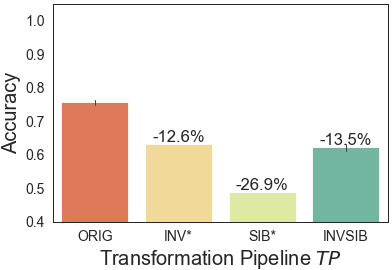}
        \caption{Yahoo! Answers}
        \label{fig:test_acc_yahoo_answers_topics}
    \end{subfigure}
    \begin{subfigure}[b]{0.23\textwidth}
        \centering
        \includegraphics[width=\textwidth]{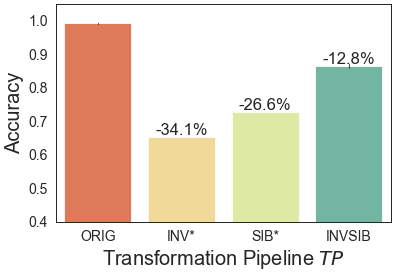}
        \caption{IMDB}
        \label{fig:test_acc_imdb}
    \end{subfigure}
    \\
    \begin{subfigure}[b]{0.23\textwidth}
        \centering
        \includegraphics[width=\textwidth]{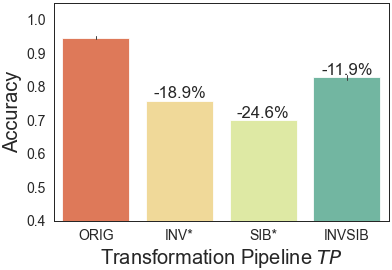}
        \caption{Amazon Polarity}
        \label{fig:test_acc_yahoo_amazon_polarity}
    \end{subfigure}
    \begin{subfigure}[b]{0.23\textwidth}
        \centering
        \includegraphics[width=\textwidth]{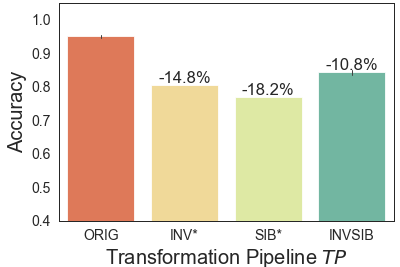}
        \caption{Yelp Polarity}
        \label{fig:test_acc_yelp_polarity}
    \end{subfigure}
    \vspace{-0.5em}
    \caption{RQ2 defect detection comparison. Percentages show change in accuracy relative to $T_{\texttt{ORIG}}$. Lower accuracy indicates greater efficacy at inducing error.}
    \label{fig:test_acc}
    \vspace{-2em}
\end{figure}

Figure \ref{fig:test_acc} shows how defect detection is enabled by INV and SIB. With the exception of Yahoo! Answers, the models scored nearly perfect accuracy on $T_{\text{ORIG}}$; however, when the same models are tested using data generated with INV and SIB, they struggle to generalize. Test data synthesized with SIB can reveal most defects in these models, indicating the value of {\em sibylvariance} in constructing test oracles for ML models in the absence of expensive human labeling and judgements.

Tests which lie outside the expected input distribution are not likely to be fair nor actionable. Since SIB transforms generally perturb data more aggressively than INV ones, they likewise possess more potential for creating unreasonable, out-of-domain tests of model quality. However, the positive results in \textbf{RQ1} may justify the use of SIB transformations as reasonable for testing. Had the newly transformed data truly belonged to a different distribution, model performance on the in-domain test set should have decreased as a result of dataset shift \cite{data_shift1, data_shift3}. In fact, we observed the opposite as model performance was consistently improved. This suggests that SIB transforms yield data that is tenably in-domain and therefore may complement INV transforms in exposing defective model behavior.

% \mknote{is the term in-domain or out-of-domain a self-explainable term without citations? [FHC: I think so]}

% \mknote{isn't this too strong? I thought our goal was to tone it down? [FHC: I think we have a strong argument that the tests are in-domain and therefore valid for testing. I didn't want to weaken the paper by undercutting this RQ.]}

We theorize that the effectiveness of SIB-generated tests comes from the expanded objectives it permits. For example, $T_{\text{TextMix}}$ assess whether the model can recognize which classes are present and to what degree. $T_{\text{SentMix}}$ does the same but further scrutinizes long-range comprehension by broadly distributing related topic sentences. Datasets with lengthy inputs are particularly vulnerable to transformations of this kind. Lastly, $T_{\text{WordMix}}$ forces the model to forgo reliance on text structure and evaluates keyword comprehension amidst noisy contexts. In contrast, most INV transformations involve minor changes --- e.g. expand contractions --- and test the aspect of language already well modeled from extensive pre-training. The INV C2S transform is an exception that drastically alters input and thus reveals more defects than other $T_{\text{INV}}$ pipelines. 

%\vspace{-1em}

\vspace{1em}
\begin{tcolorbox}
    \textbf{Defect Detection.} Models tested with SIB-transformed data exhibited the greatest number of defects in 83\% (5 out of 6) of experimental configurations. 
\end{tcolorbox}
 
\subsection{RQ3. Adversarial Robustness}
\label{RQ3}

For RQ3, we assess whether models trained on INV or SIB are more resilient to adversarial attacks than models trained on an original data. An adversarial text input is typically obtained via semantic preserving (i.e. \emph{invariant}) perturbations to legitimate examples in order to deteriorate the model performance. The changes are typically generated by ascending the gradient of the loss surface with respect to the original example and improving robustness to adversarial attacks is a necessary precondition for real-world NLP deployment.

We select three attack algorithms based on their popularity and effectiveness: (1) TextFooler \cite{textfooler}, (2) DeepWordBug \cite{deepwordbug}, and (3) TextBugger \cite{textbugger}, all as implemented in TextAttack \cite{textattack}. We focus on models trained with 10 examples per class because the largest changes in generalization performance are more likely to exhibit the clearest trend for adversarial robustness. For each of 11 models and 3 attacks, we randomly sample 100 inputs from the original data and perturb them to create a total of 3,300 adversarial examples. 

Table \ref{tab:adv_robustness} shows that, of all the cases where adversarial robustness is improved over $T_{\texttt{ORIG}}$, 92\% of them involve SIB. On average, SIB*-trained models improve robustness by 4\%, while INV*-trained models sustain a 1\% decrease. Topic classification is made more robust via training with augmented data. Consistently, $T_{\texttt{$\alpha$-SentMix}}$ produces the most resilient models. For sentiment analysis, improved generalization performance enabled by SIB does not necessarily lead to improved robustness to existing adversarial attacks.
The underlying sentiment models trained with augmented data improves generalization over $T_{\texttt{ORIG}}$ by an average of 5\%. However, counter-intuitively,
  the models are not more robust to the three attacks than $T_{\texttt{ORIG}}$ and that Pearson correlation is -0.28 between accuracy and adversarial robustness. This finding motivates future work to investigate why there is a negative correlation and how to design SIB such that accuracy improvement also translates to corresponding adversarial robustness. 
  
% \mknote{I suggest we also pick which transformation is the most robust to each adversarial attack algorithmm and present the delta: 
% Then for all six data sets, we can only pick the top performing ones  and just name the transformation (such as TexMix) that is responsible for the top one, and how much difference it has from ORG or INV. So the message that I want to send is that MIX transformation tends to be a top performing one, but this is not always consistent across different data set and that when accuracy is improved, adversarial robustness is not improved together.}

% \mknote{describe your key take away msg that more future work is necessary to investigate why adversarial robustness is not improved, when accuracy is improved. put a positive spin}

\begin{table*}[htbp]
    \vspace{-1em}
    \centering
    \resizebox{\textwidth}{!}{%
    \begin{tabular}{|c|c|ccc|c|c|ccc|c|c|ccc|}
    \hline
    \rowcolor[HTML]{EFEFEF} 
    \cellcolor[HTML]{EFEFEF}                                                             & \cellcolor[HTML]{EFEFEF}                              & \multicolumn{3}{c|}{\cellcolor[HTML]{EFEFEF}\textbf{Attack Success Rate}}                                                             & \cellcolor[HTML]{EFEFEF}                                                           & \cellcolor[HTML]{EFEFEF}                              & \multicolumn{3}{c|}{\cellcolor[HTML]{EFEFEF}\textbf{Attack Success Rate}}                                                             & \cellcolor[HTML]{EFEFEF}                                                             & \cellcolor[HTML]{EFEFEF}                              & \multicolumn{3}{c|}{\cellcolor[HTML]{EFEFEF}\textbf{Attack Success Rate}}                                                             \\ \cline{3-5} \cline{8-10} \cline{13-15} 
    \rowcolor[HTML]{EFEFEF} 
    \multirow{-2}{*}{\cellcolor[HTML]{EFEFEF}\textbf{Dataset}}                           & \multirow{-2}{*}{\cellcolor[HTML]{EFEFEF}\textbf{TP}} & \multicolumn{1}{c|}{\cellcolor[HTML]{EFEFEF}\textbf{TF}} & \multicolumn{1}{c|}{\cellcolor[HTML]{EFEFEF}\textbf{DWB}} & \textbf{TB}   & \multirow{-2}{*}{\cellcolor[HTML]{EFEFEF}\textbf{Dataset}}                         & \multirow{-2}{*}{\cellcolor[HTML]{EFEFEF}\textbf{TP}} & \multicolumn{1}{c|}{\cellcolor[HTML]{EFEFEF}\textbf{TF}} & \multicolumn{1}{c|}{\cellcolor[HTML]{EFEFEF}\textbf{DWB}} & \textbf{TB}   & \multirow{-2}{*}{\cellcolor[HTML]{EFEFEF}\textbf{Dataset}}                           & \multirow{-2}{*}{\cellcolor[HTML]{EFEFEF}\textbf{TP}} & \multicolumn{1}{c|}{\cellcolor[HTML]{EFEFEF}\textbf{TF}} & \multicolumn{1}{c|}{\cellcolor[HTML]{EFEFEF}\textbf{DWB}} & \textbf{TB}   \\ \hline
    \cellcolor[HTML]{FFFFFF}                                                             & ORIG                                                  & \multicolumn{1}{c|}{0.69}                                & \multicolumn{1}{c|}{0.56}                                 & 0.54          & \cellcolor[HTML]{FFFFFF}                                                           & ORIG                                                  & \multicolumn{1}{c|}{0.92}                                & \multicolumn{1}{c|}{0.55}                                 & 0.64          &                                                                                      & ORIG                                                  & \multicolumn{1}{c|}{0.54}                                & \multicolumn{1}{c|}{0.46}                                 & 0.52          \\ \cline{2-5} \cline{7-10} \cline{12-15} 
    \cellcolor[HTML]{FFFFFF}                                                             & INV*                                                  & \multicolumn{1}{c|}{0.66}                                & \multicolumn{1}{c|}{0.56}                                 & 0.48          & \cellcolor[HTML]{FFFFFF}                                                           & INV*                                                  & \multicolumn{1}{c|}{\textbf{0.76}}                       & \multicolumn{1}{c|}{0.47}                                 & 0.48          &                                                                                      & INV*                                                  & \multicolumn{1}{c|}{0.57}                                & \multicolumn{1}{c|}{0.49}                                 & 0.49          \\ \cline{2-5} \cline{7-10} \cline{12-15} 
    \cellcolor[HTML]{FFFFFF}                                                             & SIB*                                                  & \multicolumn{1}{c|}{\textbf{0.60}}                       & \multicolumn{1}{c|}{\textbf{0.43}}                        & \textbf{0.45} & \cellcolor[HTML]{FFFFFF}                                                           & SIB*                                                  & \multicolumn{1}{c|}{0.77}                                & \multicolumn{1}{c|}{\textbf{0.40}}                        & \textbf{0.41} &                                                                                      & SIB*                                                  & \multicolumn{1}{c|}{\textbf{0.48}}                       & \multicolumn{1}{c|}{\textbf{0.41}}                        & 0.49          \\ \cline{2-5} \cline{7-10} \cline{12-15} 
    \multirow{-4}{*}{\cellcolor[HTML]{FFFFFF}\textbf{AG News}}                           & INVSIB                                                & \multicolumn{1}{c|}{0.78}                                & \multicolumn{1}{c|}{0.62}                                 & 0.57          & \multirow{-4}{*}{\cellcolor[HTML]{FFFFFF}\textbf{DBpedia}}                         & INVSIB                                                & \multicolumn{1}{c|}{0.83}                                & \multicolumn{1}{c|}{0.56}                                 & 0.52          & \multirow{-4}{*}{\textbf{\begin{tabular}[c]{@{}c@{}}Yahoo! \\ Answers\end{tabular}}} & INVSIB                                                & \multicolumn{1}{c|}{0.54}                                & \multicolumn{1}{c|}{0.44}                                 & \textbf{0.46} \\ \hline
                                                                                         & ORIG                                                  & \multicolumn{1}{c|}{\textbf{0.48}}                       & \multicolumn{1}{c|}{0.40}                                 & 0.42          &                                                                                    & ORIG                                                  & \multicolumn{1}{c|}{\textbf{0.48}}                       & \multicolumn{1}{c|}{\textbf{0.20}}                        & \textbf{0.28} &                                                                                      & ORIG                                                  & \multicolumn{1}{c|}{0.86}                                & \multicolumn{1}{c|}{\textbf{0.25}}                        & 0.71          \\ \cline{2-5} \cline{7-10} \cline{12-15} 
                                                                                         & INV*                                                  & \multicolumn{1}{c|}{0.49}                                & \multicolumn{1}{c|}{0.42}                                 & \textbf{0.36} &                                                                                    & INV                                                   & \multicolumn{1}{c|}{0.64}                                & \multicolumn{1}{c|}{0.41}                                 & 0.52          &                                                                                      & INV*                                                  & \multicolumn{1}{c|}{0.70}                                & \multicolumn{1}{c|}{0.50}                                 & 0.68          \\ \cline{2-5} \cline{7-10} \cline{12-15} 
                                                                                         & SIB*                                                  & \multicolumn{1}{c|}{0.55}                                & \multicolumn{1}{c|}{\textbf{0.39}}                        & 0.46          &                                                                                    & SIB                                                   & \multicolumn{1}{c|}{0.61}                                & \multicolumn{1}{c|}{0.39}                                 & 0.53          &                                                                                      & SIB*                                                  & \multicolumn{1}{c|}{\textbf{0.56}}                       & \multicolumn{1}{c|}{0.32}                                 & \textbf{0.55} \\ \cline{2-5} \cline{7-10} \cline{12-15} 
    \multirow{-4}{*}{\textbf{\begin{tabular}[c]{@{}c@{}}Amazon\\ Polarity\end{tabular}}} & INVSIB                                                & \multicolumn{1}{c|}{0.65}                                & \multicolumn{1}{c|}{0.58}                                 & 0.60          & \multirow{-4}{*}{\textbf{\begin{tabular}[c]{@{}c@{}}Yelp\\ Polarity\end{tabular}}} & INVSIB                                                & \multicolumn{1}{c|}{0.75}                                & \multicolumn{1}{c|}{0.51}                                 & 0.61          & \multirow{-4}{*}{\textbf{IMDB}}                                                      & INVSIB                                                & \multicolumn{1}{c|}{0.89}                                & \multicolumn{1}{c|}{0.79}                                 & 0.88          \\ \hline
    \end{tabular}%
    }
    \vspace{-0.5em}
    \caption{RQ3 adversarial robustness comparison for INV*, SIB*, and INVSIB using TextFooler (TF), DeepWordBug (DWB), and TextBugger (TB). A lower attack success rate indicates a higher adversarial robustness.}
    \label{tab:adv_robustness}
    \vspace{-1.25em}
\end{table*}

% \begin{tcolorbox}
%     \textbf{Adversarial Robustness.} \\ Models trained upon SIB-augmented data improved adversarial robustness $11\times$ more than those trained on INV-augmented data. 
% \end{tcolorbox}

\begin{tcolorbox}
    \textbf{Adversarial Robustness.} Of all the experimental configurations where adversarial robustness was improved over the no-transform baseline, 92\% (11 out of 12) of them involved models trained on SIB-augmented data.
\end{tcolorbox}

\section{Discussion}
\label{discussion}

\paragraph{\textbf{How does sibylvariance help?}} The primary purpose of data transformations in ML is to \emph{diversify} datasets in the neighborhood of existing points, a principle formalized as Vicinal Risk Minimization (VRM) \cite{VRM}. Synthetic examples can be drawn from a vicinal distribution to find similar but different points that enlarge the original data distribution. For instance, within image classification, it is common to define the vicinity of an image as the set of its random crops, axal reflections, and other label-preserving INV transforms. While VRM can expose ML models to more diverse input space and consequently reduce generalization errors, the neighborhoods created by INV are relatively restricted. This is due to the label-preserving constraint limiting the degree of perturbation freedom on the original data. 

\begin{figure}[htbp]
    \vspace{-0.5em}
    \centering
    \begin{subfigure}[b]{0.22\textwidth}
        \centering
        \includegraphics[width=\textwidth]{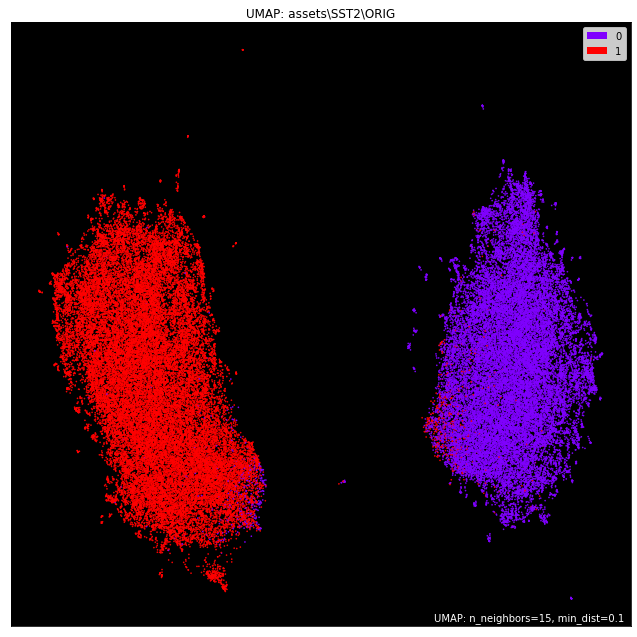}
        \caption{$T_{\text{ORIG}}$}
        \label{fig:umap_SST2_ORIG}
    \end{subfigure}
    \hfill
    \begin{subfigure}[b]{0.22\textwidth}
        \centering
        \includegraphics[width=\textwidth]{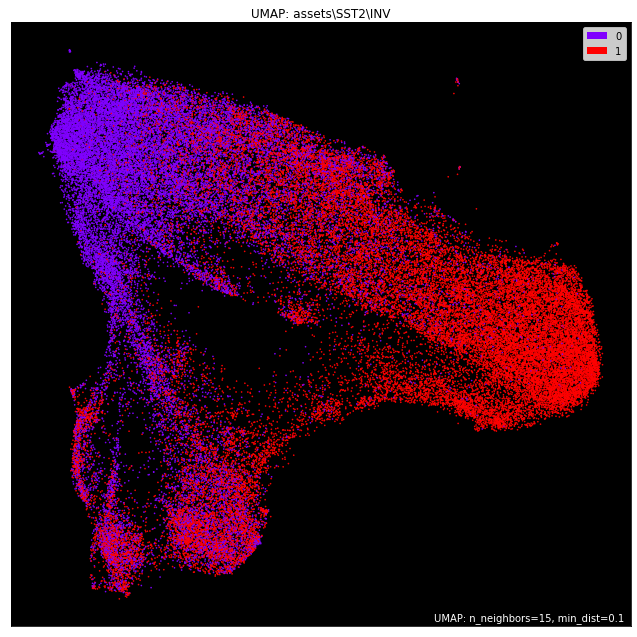}
        \caption{$T_{\text{INV}}$}
        \label{fig:umap_SST2_INV}
    \end{subfigure}
    \hfill
    \begin{subfigure}[b]{0.22\textwidth}
        \centering
        \includegraphics[width=\textwidth]{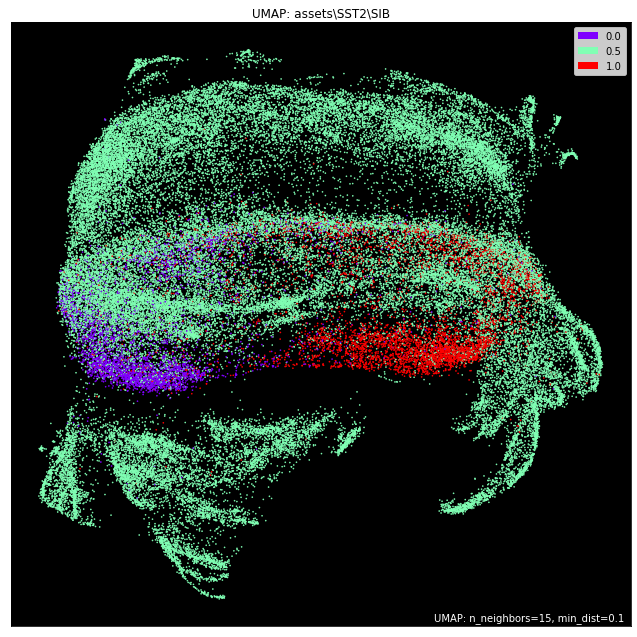}
        \caption{$T_{\text{SIB}}$}
        \label{fig:umap_SST2_SIB}
    \end{subfigure}
    \hfill
    \begin{subfigure}[b]{0.22\textwidth}
        \centering
        \includegraphics[width=\textwidth]{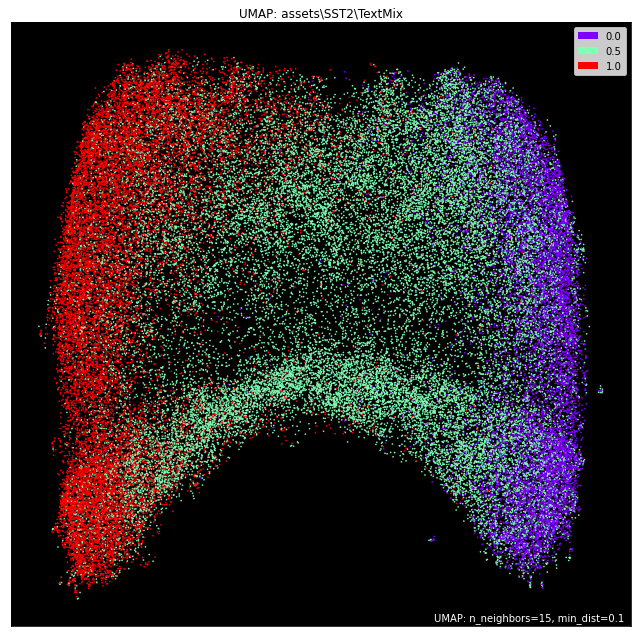}
        \caption{$T_{\text{TextMix}}$}
        \label{fig:umap_SST2_TextMix}
    \end{subfigure}
    \vspace{-0.5em}
    \caption{UMAP visualizations of BERT \texttt{[CLS]} tokens for SST-2. Blue, red, and green represent ``Negative,'' ``Positive,'' and ``Mixed'', respectively. }
    \label{fig:umap_viz}
    \vspace{-1em}
\end{figure}

SIB effectively expands the vicinity relation via transmutations and mixture mutations. Newly created data can claim full or mixed membership in  target classes. %This further diversifies the dataset and consequently promotes better generalization than the more limited INV transformations. While more theoretical analysis is required, we hypothesize that sibylvariance benefits model training in two related ways. The first is the increased input space coverage already shown to be helpful via VRM. The second is that SIB transformations may potentially maximize the margins between a model's decision surface and the different class clusters. 
To support our intuition, we visualize the effects of various transformations on SST-2~\cite{sst2}. Figure \ref{fig:umap_viz} presents the UMAP-reduced~\cite{UMAP} \texttt{[CLS]} tokens produced by a BERT transformer for sentiment classification. Figure \ref{fig:umap_SST2_ORIG} shows that the classes are initially well separated and high performance can be obtained by selecting any separating surface between the two clusters. However, a more reasonable choice for the best boundary is one that exhibits the largest margin between classes --- the very intuition behind Support Vector Machines \cite{SVM}. Figure \ref{fig:umap_SST2_TextMix} suggests that a model trained on mixture mutations is likely to arrive at a boundary with the lowest loss. For example, in \ref{fig:umap_SST2_TextMix}, the augmented examples in green provide additional loss feedback from uncovered portions of the input space to encourage a decision boundary that maximizes the margin between class clusters. A similar expectation may hold for SIB in Figure \ref{fig:umap_SST2_SIB}. However, the effects of INV transforms shown in Figure~\ref{fig:umap_SST2_INV} do not appear to support such margin maximization. 

%having additional green examples  encourages a boundary that maximizes the margin between the two closely placed clusters, challenging the learner to disambiguate better.  However, it is difficult to grasp how INV in Figure \ref{fig:umap_SST2_INV} would accomplish such margin maximization, as there exists many possible boundaries that separate blue and red.
% Instead, the baseline benefits of all the transformations might be best explained by increased input space coverage. 

\paragraph{\textbf{Threats to Validity.}} 

External threats to validity include the generalization of our results to model architectures dissimilar to BERT (i.e. \texttt{bert-base-uncased}). It is possible that larger autoencoder models like RoBERTa \cite{RoBERTa} and auto-regressive models like XLNet \cite{XLNet} may respond differently to SIB transformations. Secondly, while the framework of sibylvariance is applicable to all data types, we have only provided empirical results supporting their efficacy for text classification models. We leave the exploration of SIB applications to image, time series, and other domains to future work.

Internal threats include how we derived mixed labels for generated text. We assumed that the critical semantics can be approximated via the ratio of words contributed by source text. This assumption may not account for other linguistic interaction and thus could lead to suboptimal labels. However, SIB did significantly improve upon the INV and the ORIG baselines in the \textbf{RQ1} generalization study, suggesting that the constructed soft labels still reflected useful semantics. This indirectly supports the validity of SIB-transformed data for testing in \textbf{RQ2}, although we acknowledge that additional caution is required for using any aggressively modified, synthetic data as a substitute for real data for the purpose of exposing defective model behavior.

\section{Related Work}
\label{related_work}

In this section, we broadly cover data transformations within and outside of the text domain because our proposed framework for sibylvariance is applicable to all classification contexts. 

\vspace{-0.5em}
\paragraph{Data Augmentation.} Effective data augmentation is a key factor enabling superior model performance on a wide range of tasks \cite{alexnet, pythia, uda}. In many cases, practitioners leverage domain knowledge to reinforce critical invariances in the underlying data. In computer vision, for example, translation invariance is the idea that no matter where the objects of interest reside within an image, the model will still classify them correctly. Image translations and random crops encourage this more generalized conceptualization within the model \cite{trans_inv} and all other transforms have a similar goal: reinforce a particular invariance that helps the learner perform well on future unseen data. 

Numerous techniques have been proposed to assist with this learning objective and thereby improve generalization. Random erasing \cite{random_erasing, cutout} and noise injection \cite{time_series_aug_survey, uda} support invariance to occlusions and promote robust features. Interpolating \cite{SMOTE} and extrapolating \cite{devries2017dataset} nearest neighbors in the input / feature space reinforces a linear relationship between the newly created data and the supervision signal while reducing class imbalance. However, nearly all of these approaches, and many others \cite{image_aug_survey, nlp_aug_survey}, are label-preserving and therefore limited in their capacity to induce deeper learning of invariant concepts. 

Sibylvariant transforms enjoy several desirable aspects of INV transformations while mitigating their drawbacks. Similar to feature space functions \cite{devries2017dataset}, mixture mutations do not require significant domain knowledge. Like approaches that reduce dataset imbalance \cite{SMOTE}, SIB transforms can increase class representation through mixed membership or targeted transmutations that inherit diverse characteristics of the source inputs. In all cases, relaxing the label-preserving constraint enables SIB functions to both complement and enhance the learning of critical invariances by further expanding the support of the dataset in new directions. 

\vspace{-0.5em}
\paragraph{Adversarial Attacks \& Robustness.} Adversarial attacks are a special class of INV transformations that simultaneously minimize perturbations to the input while maximizing the perception of change to a learner. This task is more difficult within the NLP domain due to the discrete nature of text, but several works \cite{NLP_adversaries, NLP_adversaries_survey} have proven successful at inducing model errors. Real-world use of NLP requires resilience to such attacks and our work complements robust training \cite{robust_classifiers} and robust certification \cite{robust_cert, robustness_fairness} to produce more reliable models.

\vspace{-0.5em}
\paragraph{Emerging Sibylvariant Transforms.} Specific transformations designed to alter the expected class of an input have existed prior to this work \cite{mixup, cutmix, TextMixup2, CycleGAN}, albeit primarily in the image domain and also in a more isolated, ad hoc fashion. Among our primary contributions is to propose a unifying name, framework, and taxonomy for this family of sibylvariant functions. Furthermore, most prior works introduce a single transformation and evaluate its efficacy on training alone. In contrast, we proposed several novel transformations, a new adaptive training routine, and evaluated the broader impacts of 41 INV and SIB transforms on training, defect detection, and  robustness simultaneously. 

Recently published examples of SIB mixture mutations for text \cite{TextMixup, mixtext} differ from ours in several important ways. Prior work operates exclusively within the \emph{hidden} space inside specific models, which limits transferability between different algorithm types. All of our transformations operate in the \emph{input} space, which is both more general and more challenging because we have to contend with rules of grammar and style. However, this also provides greater transparency. Furthermore, because our overall approach samples from 41 different transformations, we are able to exercise a broader range of model behaviors. For example, \texttt{SentMix} is designed to encourage long-range understanding, while other transforms evoke their own specific objectives. Any individual transformation is inherently more limited, e.g. TMix can only encourage the model to behave linearly for borderline cases.

\vspace{-0.25em}

\section{Conclusion}
\label{conclusion}

\vspace{-0.25em}

Inspired by metamorphic testing, we proposed the notion of \emph{sibylvariance} to jointly transform both input and output class $(X_i,y_i)$ pairs in a knowable way. To explore the potential of sibylvariance, we define 18 new text transformations and adapt 23 existing transformations into an open source tool called {\tool}. In particular, we define several types of mixture mutations and design a novel concept-based text transformation technique utilizing salience attribution and neural sentence generation. Across six benchmarks from two different NLP classification tasks, we systematically assess the effectiveness of INV and SIB for generalization performance, defect detection, and adversarial robustness. Our extensive evaluation shows that many SIB transforms, and especially the {\em adaptive} mixture mutations, are extremely effective. SIB achieves the highest training accuracy in 89\% of the experimental configurations. When used for testing, SIB test suites reveal the greatest number of model defects in 5 out of 6 benchmarks. Finally, models trained on SIB-augmented data improve adversarial robustness $11\times$ more often than those trained on INV-augmented data. 

% The comprehensive collection of text transformations in 7 categories, spanning from mixture, word swap, negation, emojis, punctuation, text insertion and punctuation are packaged as an open source tool, called {\tool}. 

% We have proposed the notion of \emph{sibylvariance} to jointly transform $(X_i,y_i)$ pairs for classification tasks in a knowable way. We explored the effectiveness of both INV and SIB transformations for testing and training a variety of models in the image and text domains, although sibylvariance itself is domain agnostic and widely applicable. Through an extensive evaluation, we have shown that SIB transforms --- especially the mixture mutations --- are generally more effective than INV transforms and can significantly improve the creation of defect detecting tests and the generalization performance during training. Lastly, we have packaged many of the existing and newly proposed transformations in an open source tool called \texttt{Sibyl}. 

% Inspired by the conceptual framework of sibylvariance, one potential mixture mutation would be to automatically generate new, semantically intelligible terms via hyphenation or as portmanteaus. Existing words like ``smog'' and ``frenemy'' combine two different concepts into a single token whose meaning can be understood via its parts and it may be worthwhile to generate new out-of-vocabulary (OOV) terms via a tool like Entendrepreneur \citep{entendrepreneur} to determine if language models can comprehend the component meanings \citep{blendtest}.

% \miryung{quantify}
% \miryung{how much, quantify?}
\vspace{-0.25em}

\section*{Acknowledgements}
\label{acknowledgements}

\vspace{-0.25em}

This work is supported in part by National Science Foundations via grants CCF-2106420, CCF-2106404, CNS-2106838, CCF-1764077, CHS-1956322, CCF-1723773, ONR grant N00014-18-1-2037, Intel CAPA grant, Samsung, and a CISCO research contract. We would also like to thank Atharv Sakhala for early contributions to the {\tool} project as well as Jason Teoh, Sidi Lu, Aaron Hatrick, Sean Gildersleeve, Hannah Pierce, and all the anonymous reviewers for their many helpful suggestions.

\bibliography{refs}
\bibliographystyle{acl_natbib}

\clearpage

%%%%%%%%%%%%%%%%%%%%%%%%%%%%%%%%%%%%%%%%%%%%%%%%%%%%%%%%%%%%

\appendix

\begin{table*}[t]
    \section{Implemented {\tool} Transformations}
    \label{appendix: transformations}
    \centering
    \begin{tabular}{l l c c}
        \toprule
        \textbf{Category} & \textbf{Transformation} & \textbf{Sentiment} & \textbf{Topic} \\
        \midrule
        % Cateogry 1 : Mixture
        Mixture     & TextMix                               & SIB  & SIB \\
        Mixture     & SentMix                               & SIB  & SIB \\
        Mixture     & WordMix                               & SIB  & SIB \\
        \midrule
        % Category 2: Generative
        Generative  & Concept2Sentence                      & INV  & INV \\
        % Generative  & Concept2Sentence (antonymize)         & SIB  & INV \\
        % Generative  & Concept2Sentence (synonymize)         & INV  & INV \\
        Generative  & ConceptMix                            & SIB  & SIB \\
        \midrule
        % Category 3 : Word Swap 
        Word Swap   & replace antonym                       & SIB  & INV \\
        Word Swap   & replace cohyponym                     & INV  & INV \\
        Word Swap   & replace hypernym                      & INV  & INV \\
        Word Swap   & replace hyponym                       & INV  & INV \\
        Word Swap   & replace synonym (wordnet)             & INV  & INV \\
        Word Swap   & change numbers (except 2 and 4)       & INV* & INV \\
        Word Swap   & change locations based on dictionary  & INV  & INV \\
        Word Swap   & change names based on dictionary      & INV  & INV \\
        \midrule
        % Cateogry 4: Negation
        Negation    & add negation                          & INV* & INV \\
        Negation    & remove negation                       & INV* & INV \\
        \midrule
        % Category 5: punctuation %
        Punctuation & expand contractions                   & INV  & INV \\
        Punctuation & reduce contractions                   & INV  & INV \\
        \midrule
        % Category 6: Text insertion 
        Text Insertion & add URL to negative content        & SIB  & INV \\
        Text Insertion & add URL to positive content        & SIB  & INV \\
        Text Insertion & add negative phrase                & SIB  & INV \\
        Text Insertion & add positive phrase                & SIB  & INV \\
        \midrule
        % Category 7 Typo 
        Typos       & char deletion                         & INV* & INV \\
        Typos       & char insertion                        & INV* & INV \\
        Typos       & char movement (n spaces)              & INV* & INV \\
        Typos       & char repacement (homoglyph)           & INV  & INV \\
        Typos       & char replacement                      & INV* & INV \\
        Typos       & char swap (n spaces)                  & INV* & INV \\
        Typos       & char swap (QWERTY)                    & INV* & INV \\
        Typos       & word deletion                         & INV* & INV \\
        Typos       & word insertion                        & INV* & INV \\
        Typos       & word replacement                      & INV* & INV \\
        Typos       & word replacement (homophone)          & INV  & INV \\
        Typos       & word swap                             & INV* & INV \\
        \midrule
        %category 8: Emoji
        Emojis      & replace words with emojis (Emojify)   & INV  & INV \\
        Emojis      & replace emojis with words (Demojify)  & INV  & INV \\
        Emojis      & add negative emoji                    & SIB  & INV \\
        Emojis      & add neutral emoji                     & INV  & INV \\
        Emojis      & add positive emoji                    & SIB  & INV \\
        Emojis      & remove negative emoji                 & SIB  & INV \\
        Emojis      & remove neutral emoji                  & INV  & INV \\
        Emojis      & remove positive emoji                 & SIB  & INV \\
        \bottomrule
    \end{tabular}
    \caption{Transform descriptions currently implemented in {\tool}, sampled from according to task (sentiment analysis or topic) and $TP$ (INV, SIB, or INVSIB). Note that transformations are INV or SIB with respect to specific tasks. Asterisks (*) indicate that the variance type could be either INV or SIB, but the listed variance was judged to be more likely.}
    \label{tbl: transformations_appendix}
\end{table*}

\begin{table*}[t]
    \section{Other Possible Text Transformations}
    \label{appendix: other_transformations}
    \centering
    \begin{tabular}{l l c c}
        \toprule
        \textbf{Category} & \textbf{Transformation} & \textbf{Sentiment} & \textbf{Topic} \\
        Word Swap      & replace synonym (embedding)           & INV  & INV  \\
        Word Swap      & word swap (masked)                    & INV* & INV  \\
        Word Swap      & change gendered pronoun               & INV  & INV* \\
        Word Swap      & change protected class                & INV  & INV* \\
        Word Swap      & change "for" to 4                     & INV  & INV  \\
        Word Swap      & change "to" to 2                      & INV  & INV  \\
        Word Swap      & swap phrase with acronym              & INV  & INV  \\
        \midrule
        Negation       & negation of negative clause           & SIB  & INV  \\
        Negation       & negation of neutral clause            & INV  & INV  \\
        Negation       & negation of positive clause           & SIB  & INV  \\
        \midrule
        Paraphrase     & backtranslation                       & INV  & INV  \\
        \midrule
        Punctuation    & add exclamation                       & INV* & INV  \\
        Punctuation    & add period                            & INV  & INV  \\
        Punctuation    & add question mark                     & INV  & INV  \\
        Punctuation    & remove exclamation                    & SIB* & INV  \\
        Punctuation    & remove period                         & INV  & INV  \\
        Punctuation    & remove question mark                  & INV  & INV  \\
        \midrule
        Text Insertion & add random URL (404)                  & INV  & INV  \\
        Text Insertion & add neutral phrase                    & INV  & INV  \\
        \midrule
        Tense / Voice  & make continuous future tense          & INV* & INV  \\
        Tense / Voice  & make continuous past tense            & INV* & INV  \\
        Tense / Voice  & make continuous present tense         & INV* & INV  \\
        Tense / Voice  & make perfect continuous future tense  & INV* & INV  \\
        Tense / Voice  & make perfect continuous past tense    & INV* & INV  \\
        Tense / Voice  & make perfect continuous present tense & INV* & INV  \\
        Tense / Voice  & make perfect future tense             & INV* & INV  \\
        Tense / Voice  & make perfect past tense               & INV* & INV  \\
        Tense / Voice  & make perfect present tense            & INV* & INV  \\
        Tense / Voice  & make simple future tense              & INV* & INV  \\
        Tense / Voice  & make simple past tense                & INV* & INV  \\
        Tense / Voice  & make simple present tense             & INV* & INV  \\
        Tense / Voice  & change voice active                   & INV  & INV  \\
        Tense / Voice  & change voice passive                  & INV  & INV  \\
        \midrule
        Emojis         & replace emoji with word antonym       & SIB  & INV  \\
        Emojis         & replace emoji with word synonym       & INV  & INV  \\
        Emojis         & replace word with emoji antonym       & SIB  & INV  \\
        Emojis         & replace word with emoji synonym       & INV  & INV  \\
        \bottomrule
    \end{tabular}
    \caption{Transform NOT currently implemented in {\tool}, but represent potentially interesting directions for future work. Asterisks (*) indicate that the variance type could be either INV or SIB, but the listed variance was judged to be more likely.}
    \label{tbl: transformations_unimplemented_appendix}
\end{table*}

\clearpage

\begin{table*}[htbp]
\section{Sibylvariant Subtype Examples}
\label{appendix: sib_subtypes}
\centering
\begin{tabular}{lll}
    \textbf{SIB Subtype} &
      \multicolumn{1}{c}{\textbf{\begin{tabular}[c]{@{}c@{}}Image\\ (Classification)\end{tabular}}} &
      \multicolumn{1}{c}{\textbf{\begin{tabular}[c]{@{}c@{}}Text\\ (Sentiment Analysis)\end{tabular}}} \\
    \toprule
    \begin{tabular}[l]{@{}p{97pt}@{}}
        \textbf{Transmutation} \\ 
        $A \rightarrow B$ \\ 
        (Hard Label) \\ \\ 
        Changes one class into another class, while retaining stylistic elements of the original. \\
    \end{tabular} &
    \begin{tabular}[l]{@{}C{150pt}@{}}
        \textbf{\emph{Rotation}} \\ 
        \includegraphics[width=0.29\textwidth]{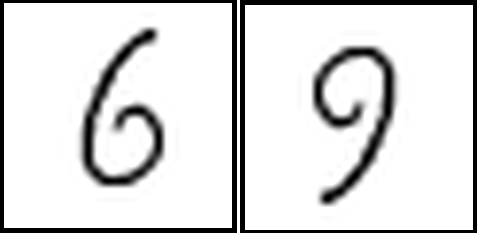} \\ 
        Digit 6 $\rightarrow$ Digit 9 \\ \\
        
        \textbf{\emph{GAN-based Object Transfiguration}} \\ 
        \includegraphics{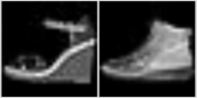} \\ 
        Sandal $\rightarrow$ Sneaker \\ 
    \end{tabular} &
    \begin{tabular}[l]{@{}C{150pt}@{}}
        \textbf{\emph{Antonym Replacement}} \\ \textcolor{PineGreen}{I love NY} \\ $\downarrow$ \\ \textcolor{RedOrange}{I \underline{hate} NY} \\ \\ 
        \textbf{\emph{Clause Negation}} \\ \textcolor{PineGreen}{You are a good person.} \\ $\downarrow$ \\ \textcolor{RedOrange}{You are \underline{not} a good person.} \\ \\
        \textbf{\emph{Stock Phrase Insertion}} \\ \textcolor{PineGreen}{It was a clever movie.} \\ $\downarrow$ \\ \textcolor{RedOrange}{It was a clever movie.} \\ \textcolor{RedOrange}{\underline{That said, I absolutely hated it.}} \\
    \end{tabular} \\
    \midrule
    \begin{tabular}[l]{@{}p{97pt}@{}}
        \textbf{Mixture Mutation} \\ 
        $A + B \rightarrow AB$ \\ 
        (Soft Label) \\ \\ 
        Mixes two or more class labels into a single data point and then interpolates the expected behavior. 
    \end{tabular} &
    \begin{tabular}[l]{@{}C{150pt}@{}}
        \textbf{\emph{Mixup}} \cite{mixup} \hspace{15pt} \textbf{\emph{Cutmix}} \cite{cutmix} \\
        \includegraphics[width=0.15\textwidth]{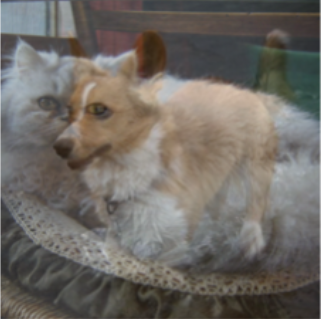} 
        \includegraphics[width=0.15\textwidth]{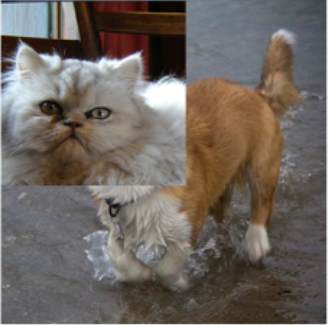} \\
        $[1, 0] + [0, 1] \rightarrow [0.35, 0.65]$ \\ \\
        \textbf{\emph{Tile}} \\
        \includegraphics[width=0.33\textwidth]{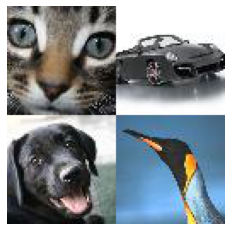} \\
        $[1, 0, 0, 0] + [0, 0, 1, 0] \; + \: $ \\ $[0, 1, 0, 0] + [0, 0, 0, 1] \rightarrow $ \\ $ [0.25, 0.25, 0.25, 0.25]$ \\ \\
    \end{tabular} &
    \begin{tabular}[l]{@{}C{150pt}@{}}
        \textbf{\emph{TextMix}} \\
        \textcolor{RedOrange}{virutally unwatchable...} \\ $+$ \\ \textcolor{PineGreen}{a vivid, thoughtful, unapologetically raw coming-of-age tale full of sex, drugs and rock 'n' roll.} \\ $=$ \\
        \textcolor{RedOrange}{virutally unwatchable...} \textcolor{PineGreen}{a vivid, thoughtful, unapologetically raw coming-of-age tale full of sex, drugs and rock 'n' roll.} \\
        $[1, 0] + [0, 1] \rightarrow [0.17, 0.83]$ \\ \\
        \textbf{\emph{WordMix}} \\
        \textcolor{RedOrange}{it is essentially empty} \\ $+$ \\ \textcolor{PineGreen}{this is a visually stunning} \\ \textcolor{PineGreen}{rumination on love} \\ $=$ \\
        \textcolor{PineGreen}{love visually} \textcolor{RedOrange}{is} \textcolor{PineGreen}{is} \textcolor{RedOrange}{essentially} \textcolor{PineGreen}{rumination on} \textcolor{RedOrange}{it} \textcolor{PineGreen}{stunning this a} \textcolor{RedOrange}{empty}
        \\
        $[1, 0] + [0, 1] \rightarrow [0.33, 0.67]$ \\
    \end{tabular} \\
    \bottomrule
    \\
    \end{tabular} 
    \caption{Examples of SIB transformations for the image and text domains. For mixture mutations, we show a soft label proportional to the pixel and word counts of their constituent parts.}
    \label{tab:sib_subtypes}
\end{table*}

\clearpage

\begin{table*}[htbp]
\section{RQ1. Detailed Training Results}
\centering
\resizebox{\textwidth}{!}{%
\begin{tabular}{|c|c|c|c|c|c|c|c|c|c|}
\hline
\rowcolor[HTML]{EFEFEF} 
\textbf{Dataset} & \textbf{TP} & \textbf{10} & \textbf{200} & \textbf{2500} & \textbf{Dataset} & \textbf{TP} & \textbf{10} & \textbf{200} & \textbf{2500} \\ \hline
 & \cellcolor[HTML]{DE7959}ORIG & 75.08 & 88.70 & 91.65 &  & \cellcolor[HTML]{DE7959}ORIG & 67.30 & 89.22 & 92.08 \\ \cline{2-5} \cline{7-10} 
 & \cellcolor[HTML]{F0D999}INV & \textbf{84.28} & 89.46 & 91.95 &  & \cellcolor[HTML]{F0D999}INV & 71.09 & 89.53 & 92.21 \\ \cline{2-5} \cline{7-10} 
 & \cellcolor[HTML]{F0D999}C2S & 82.82 & 87.84 & 91.43 &  & \cellcolor[HTML]{F0D999}C2S & 73.69 & 86.76 & 90.20 \\ \cline{2-5} \cline{7-10} 
 & \cellcolor[HTML]{DEE9A4}SIB & 83.52 & 89.20 & 91.55 &  & \cellcolor[HTML]{DEE9A4}SIB & 69.23 & 87.00 & 91.45 \\ \cline{2-5} \cline{7-10} 
 & \cellcolor[HTML]{DEE9A4}TextMix & 83.53 & 89.17 & 91.58 &  & \cellcolor[HTML]{DEE9A4}TextMix & 68.20 & 88.63 & 91.46 \\ \cline{2-5} \cline{7-10} 
 & \cellcolor[HTML]{DEE9A4}SentMix & 83.56 & 89.28 & 91.49 &  & \cellcolor[HTML]{DEE9A4}SentMix & 71.22 & 88.85 & 91.28 \\ \cline{2-5} \cline{7-10} 
 & \cellcolor[HTML]{DEE9A4}WordMix & 82.61 & 88.59 & 90.42 &  & \cellcolor[HTML]{DEE9A4}WordMix & 60.27 & 85.40 & 87.68 \\ \cline{2-5} \cline{7-10} 
 & \cellcolor[HTML]{DEE9A4}$\alpha$TextMix & 81.53 & 89.51 & 92.20 &  & \cellcolor[HTML]{DEE9A4}$\alpha$TextMix & \textbf{74.90} & \textbf{90.03} & \textbf{92.26} \\ \cline{2-5} \cline{7-10} 
 & \cellcolor[HTML]{DEE9A4}$\alpha$SentMix & 77.28 & \textbf{89.80} & \textbf{92.42} &  & \cellcolor[HTML]{DEE9A4}$\alpha$SentMix & 64.19 & 90.01 & 92.16 \\ \cline{2-5} \cline{7-10} 
 & \cellcolor[HTML]{DEE9A4}$\alpha$WordMix & 83.13 & 89.46 & 91.91 &  & \cellcolor[HTML]{DEE9A4}$\alpha$WordMix & 64.21 & 89.09 & 91.98 \\ \cline{2-5} \cline{7-10} 
 & \cellcolor[HTML]{72B6A1}INVSIB & 84.09 & 89.00 & 91.36 &  & \cellcolor[HTML]{72B6A1}INVSIB & 73.50 & 89.06 & 91.26 \\ \cline{2-5} \cline{7-10} 
 & TMix $\ddagger$ & 81.38 & 88.62 & 89.43 &  & TMix $\ddagger$ & 62.14 & 87.98 & 91.00 \\ \cline{2-5} \cline{7-10} 
 & EDA $\ddagger$ & 81.50 & 88.98 & 90.93 &  & EDA $\ddagger$ & 59.40 & 87.68 & 92.20 \\ \cline{2-5} \cline{7-10} 
\multirow{-14}{*}{\textbf{AG News}} & AEDA $\ddagger$ & 81.03 & 88.74 & 92.09 & \multirow{-14}{*}{\textbf{\begin{tabular}[c]{@{}c@{}}Amazon\\ Polarity\end{tabular}}} & AEDA $\ddagger$ & 64.72 & 88.92 & 91.83 \\ \hline
 & \cellcolor[HTML]{DE7959}ORIG & 95.71 & 98.87 & 98.96 &  & \cellcolor[HTML]{DE7959}ORIG & 74.62 & 91.66 & 93.70 \\ \cline{2-5} \cline{7-10} 
 & \cellcolor[HTML]{F0D999}INV & 97.29 & 98.81 & 99.00 &  & \cellcolor[HTML]{F0D999}INV & 77.92 & 92.00 & 94.29 \\ \cline{2-5} \cline{7-10} 
 & \cellcolor[HTML]{F0D999}C2S & 96.23 & 98.36 & 96.41 &  & \cellcolor[HTML]{F0D999}C2S & \textbf{83.91} & 89.59 & 92.80 \\ \cline{2-5} \cline{7-10} 
 & \cellcolor[HTML]{DEE9A4}SIB & 95.26 & 98.73 & 97.60 &  & \cellcolor[HTML]{DEE9A4}SIB & 78.67 & 91.89 & 93.69 \\ \cline{2-5} \cline{7-10} 
 & \cellcolor[HTML]{DEE9A4}TextMix & \textbf{97.96} & 98.88 & 97.86 &  & \cellcolor[HTML]{DEE9A4}TextMix & 79.27 & 91.07 & 93.36 \\ \cline{2-5} \cline{7-10} 
 & \cellcolor[HTML]{DEE9A4}SentMix & 97.95 & 98.86 & 99.01 &  & \cellcolor[HTML]{DEE9A4}SentMix & 80.46 & 91.96 & 93.62 \\ \cline{2-5} \cline{7-10} 
 & \cellcolor[HTML]{DEE9A4}WordMix & 97.03 & 97.89 & 98.59 &  & \cellcolor[HTML]{DEE9A4}WordMix & 74.47 & 88.39 & 92.12 \\ \cline{2-5} \cline{7-10} 
 & \cellcolor[HTML]{DEE9A4}$\alpha$TextMix & 97.72 & 98.87 & 99.04 &  & \cellcolor[HTML]{DEE9A4}$\alpha$TextMix & 77.72 & 91.73 & 94.50 \\ \cline{2-5} \cline{7-10} 
 & \cellcolor[HTML]{DEE9A4}$\alpha$SentMix & 96.38 & \textbf{98.90} & \textbf{99.06} &  & \cellcolor[HTML]{DEE9A4}$\alpha$SentMix & 76.63 & \textbf{92.60} & \textbf{94.69} \\ \cline{2-5} \cline{7-10} 
 & \cellcolor[HTML]{DEE9A4}$\alpha$WordMix & 97.01 & \textbf{98.90} & 98.90 &  & \cellcolor[HTML]{DEE9A4}$\alpha$WordMix & 78.30 & 91.50 & 93.67 \\ \cline{2-5} \cline{7-10} 
 & \cellcolor[HTML]{72B6A1}INVSIB & 95.64 & 98.74 & 98.92 &  & \cellcolor[HTML]{72B6A1}INVSIB & 78.90 & 91.85 & 93.03 \\ \cline{2-5} \cline{7-10} 
 & TMix $\ddagger$ & 95.76 & 98.53 & 98.55 &  & TMix $\ddagger$ & 61.81 & 91.19 & 92.80 \\ \cline{2-5} \cline{7-10} 
 & EDA $\ddagger$ & 97.42 & 98.63 & 98.89 &  & EDA $\ddagger$ & 71.90 & 90.88 & 94.11 \\ \cline{2-5} \cline{7-10} 
\multirow{-14}{*}{\textbf{DBpedia}} & AEDA $\ddagger$ & 97.30 & 98.88 & 98.89 & \multirow{-14}{*}{\textbf{\begin{tabular}[c]{@{}c@{}}Yelp\\ Polarity\end{tabular}}} & AEDA $\ddagger$ & 79.39 & 91.60 & 94.06 \\ \hline
 & \cellcolor[HTML]{DE7959}ORIG & 56.24 & 69.77 & 73.18 &  & \cellcolor[HTML]{DE7959}ORIG & 64.70 & 86.96 & 90.02 \\ \cline{2-5} \cline{7-10} 
 & \cellcolor[HTML]{F0D999}INV & 60.24 & 69.21 & 72.53 &  & \cellcolor[HTML]{F0D999}INV & 76.20 & 86.94 & 89.69 \\ \cline{2-5} \cline{7-10} 
 & \cellcolor[HTML]{F0D999}C2S & 61.39 & 67.31 & 70.60 &  & \cellcolor[HTML]{F0D999}C2S & 70.18 & 85.67 & 86.98 \\ \cline{2-5} \cline{7-10} 
 & \cellcolor[HTML]{DEE9A4}SIB & 61.30 & 68.45 & 73.18 &  & \cellcolor[HTML]{DEE9A4}SIB & 73.51 & 86.38 & 88.71 \\ \cline{2-5} \cline{7-10} 
 & \cellcolor[HTML]{DEE9A4}TextMix & \textbf{62.47} & 68.72 & 72.08 &  & \cellcolor[HTML]{DEE9A4}TextMix & 73.23 & 85.24 & 89.45 \\ \cline{2-5} \cline{7-10} 
 & \cellcolor[HTML]{DEE9A4}SentMix & 60.95 & 68.72 & 72.07 &  & \cellcolor[HTML]{DEE9A4}SentMix & 76.75 & 85.55 & 89.10 \\ \cline{2-5} \cline{7-10} 
 & \cellcolor[HTML]{DEE9A4}WordMix & 59.98 & 67.66 & 72.96 &  & \cellcolor[HTML]{DEE9A4}WordMix & 67.15 & 84.19 & 88.23 \\ \cline{2-5} \cline{7-10} 
 & \cellcolor[HTML]{DEE9A4}$\alpha$TextMix & 60.26 & 69.89 & 73.15 &  & \cellcolor[HTML]{DEE9A4}$\alpha$TextMix & 74.09 & 87.52 & 90.60 \\ \cline{2-5} \cline{7-10} 
 & \cellcolor[HTML]{DEE9A4}$\alpha$SentMix & 59.10 & \textbf{70.10} & 73.00 &  & \cellcolor[HTML]{DEE9A4}$\alpha$SentMix & \textbf{79.74} & \textbf{87.65} & \textbf{90.90} \\ \cline{2-5} \cline{7-10} 
 & \cellcolor[HTML]{DEE9A4}$\alpha$WordMix & 60.74 & 69.99 & \textbf{73.37} &  & \cellcolor[HTML]{DEE9A4}$\alpha$WordMix & 73.01 & 86.92 & 87.85 \\ \cline{2-5} \cline{7-10} 
 & \cellcolor[HTML]{72B6A1}INVSIB & 62.01 & 67.75 & 73.16 &  & \cellcolor[HTML]{72B6A1}INVSIB & 75.04 & 87.04 & 88.24 \\ \cline{2-5} \cline{7-10} 
 & TMix $\ddagger$ & 53.68 & 69.03 & 69.50 &  & TMix $\ddagger$ & 62.45 & 86.94 & 88.29 \\ \cline{2-5} \cline{7-10} 
 & EDA $\ddagger$ & 57.88 & 68.03 & 69.15 &  & EDA $\ddagger$ & 67.37 & 86.45 & 89.07 \\ \cline{2-5} \cline{7-10} 
\multirow{-14}{*}{\textbf{\begin{tabular}[c]{@{}c@{}}Yahoo! \\ Answers\end{tabular}}} & AEDA $\ddagger$ & 59.51 & 67.37 & 69.91 & \multirow{-14}{*}{\textbf{IMDB}} & AEDA $\ddagger$ & 72.61 & 86.56 & 88.63 \\ \hline
\end{tabular}%
}
\caption{Performance (test set accuracy (\%)) for all $TP$s. The results are averaged across three runs. Models are trained with either 10, 200, or 2500 examples per class. $TP$s are color coded by their variant type, where orange and light green are invariant and sibylvariant, respectively. White with a $\ddagger$ indicates related works for comparison. For TMix, EDA, and AEDA, we used the author's open source code with their default / recommended configurations to transform the training datasets. However, we maintained the same model training hyperparameters as our other $TP$s to facilitate fair comparisons with our work.}
\label{tab:training_acc_full}
\end{table*}

\begin{table*}[htbp]
\centering
\begin{tabular}{|l|l|r|}
\hline
\rowcolor[HTML]{EFEFEF}
\textbf{Transform}   & \textbf{Type} & \textbf{Avg $\Delta$ (\%)} \\ \hline
$\alpha$SentMix      & SIB           & +4.26              \\ \hline
$\alpha$TextMix      & SIB           & +3.55              \\ \hline
RandomCharInsert     & INV           & +3.55              \\ \hline
TextMix              & SIB           & +3.22              \\ \hline
Concept2Sentence     & INV           & +2.70              \\ \hline
AddPositiveLink      & INV / SIB     & +2.48              \\ \hline
AddNegativeEmoji     & INV / SIB     & +2.45              \\ \hline
SentMix              & SIB           & +2.33              \\ \hline
ExpandContractions   & INV           & +2.15              \\ \hline
RandomCharSubst      & INV           & +2.06              \\ \hline
AddNeutralEmoji      & INV           & +1.90              \\ \hline
RandomInsertion      & INV           & +1.72              \\ \hline
AddNegativeLink      & INV / SIB     & +1.64              \\ \hline
$\alpha$WordMix      & SIB           & +1.62              \\ \hline
ChangeNumber         & INV           & +1.44              \\ \hline
AddPositiveEmoji     & INV / SIB     & +1.25              \\ \hline
InsertNegativePhrase & INV / SIB     & +1.15              \\ \hline
RemoveNegation       & INV           & +1.00              \\ \hline
WordDeletion         & INV           & +0.86              \\ \hline
RandomSwapQwerty     & INV           & +0.83              \\ \hline
RandomCharSwap       & INV           & +0.77              \\ \hline
ContractContractions & INV           & +0.69              \\ \hline
Emojify              & INV           & +0.59              \\ \hline
ChangeLocation       & INV           & +0.37              \\ \hline
Demojify             & INV           & +0.34              \\ \hline
AddNegation          & INV           & +0.13              \\ \hline
WordMix              & SIB           & +0.08              \\ \hline
ConceptMix           & SIB           & -0.11              \\ \hline
RandomCharDel        & INV           & -0.16              \\ \hline
RemovePositiveEmoji  & INV           & -0.24              \\ \hline
RandomSwap           & INV           & -0.28              \\ \hline
ImportLinkText       & INV           & -0.56              \\ \hline
ChangeHyponym        & INV           & -0.63              \\ \hline
RemoveNeutralEmoji   & INV           & -0.72              \\ \hline
RemoveNegativeEmoji  & INV / SIB     & -0.80              \\ \hline
ChangeName           & INV           & -0.84              \\ \hline
InsertPositivePhrase & INV / SIB     & -0.95              \\ \hline
ChangeSynonym        & INV           & -1.26              \\ \hline
ChangeHypernym       & INV           & -1.78              \\ \hline
ChangeAntonym        & INV / SIB     & -2.82              \\ \hline
HomoglyphSwap        & INV           & -3.78              \\ \hline
\end{tabular}
\caption{Performance (test set accuracy (\%)) for individual transforms over a no-transform baseline averaged across all datasets. The INV / SIB types were SIB for the sentiment analysis datasets and INV for the topic classification datasets.}
\label{tab:training_acc_by_transform}
\end{table*}

\pagebreak

\begin{table*}[htbp]
\section{RQ2. Detailed Defect Detection Results}
\centering
% \resizebox{\textwidth}{!}{%
\begin{tabular}{|c|c|c|c|c|c|}
\hline
\rowcolor[HTML]{EFEFEF} 
\multicolumn{1}{|l|}{\cellcolor[HTML]{EFEFEF}\textbf{Dataset}}                       & \textbf{TP}                     & \multicolumn{1}{l|}{\cellcolor[HTML]{EFEFEF}\textbf{Test Suite Accuracy}} & \multicolumn{1}{l|}{\cellcolor[HTML]{EFEFEF}\textbf{Dataset}}                        & \textbf{TP}                     & \multicolumn{1}{l|}{\cellcolor[HTML]{EFEFEF}\textbf{Test Suite Accuracy}} \\ \hline
                                                                                     & \cellcolor[HTML]{DE7959}ORIG    & 96.22                                                                     &                                                                                      & \cellcolor[HTML]{DE7959}ORIG    & 94.68                                                                     \\ \cline{2-3} \cline{5-6} 
                                                                                     & \cellcolor[HTML]{F0D999}INV     & 89.77                                                                     &                                                                                      & \cellcolor[HTML]{F0D999}INV     & 86.91                                                                     \\ \cline{2-3} \cline{5-6} 
                                                                                     & \cellcolor[HTML]{F0D999}C2S     & 66.67                                                                     &                                                                                      & \cellcolor[HTML]{F0D999}C2S     & 75.78                                                                     \\ \cline{2-3} \cline{5-6} 
                                                                                     & \cellcolor[HTML]{DEE9A4}SIB     & 74.77                                                                     &                                                                                      & \cellcolor[HTML]{DEE9A4}SIB     & 80.99                                                                     \\ \cline{2-3} \cline{5-6} 
                                                                                     & \cellcolor[HTML]{DEE9A4}TextMix & 59.97                                                                     &                                                                                      & \cellcolor[HTML]{DEE9A4}TextMix & 79.83                                                                     \\ \cline{2-3} \cline{5-6} 
                                                                                     & \cellcolor[HTML]{DEE9A4}SentMix & 60.48                                                                     &                                                                                      & \cellcolor[HTML]{DEE9A4}SentMix & 79.83                                                                     \\ \cline{2-3} \cline{5-6} 
                                                                                     & \cellcolor[HTML]{DEE9A4}WordMix & \textbf{58.82}                                                            &                                                                                      & \cellcolor[HTML]{DEE9A4}WordMix & \textbf{70.08}                                                            \\ \cline{2-3} \cline{5-6} 
\multirow{-8}{*}{\textbf{AG News}}                                                   & \cellcolor[HTML]{72B6A1}INVSIB  & 74.50                                                                     & \multirow{-8}{*}{\textbf{\begin{tabular}[c]{@{}c@{}}Amazon\\ Polarity\end{tabular}}} & \cellcolor[HTML]{72B6A1}INVSIB  & 82.78                                                                     \\ \hline
                                                                                     & \cellcolor[HTML]{DE7959}ORIG    & 99.04                                                                     &                                                                                      & \cellcolor[HTML]{DE7959}ORIG    & 95.15                                                                     \\ \cline{2-3} \cline{5-6} 
                                                                                     & \cellcolor[HTML]{F0D999}INV     & 93.27                                                                     &                                                                                      & \cellcolor[HTML]{F0D999}INV     & 89.76                                                                     \\ \cline{2-3} \cline{5-6} 
                                                                                     & \cellcolor[HTML]{F0D999}C2S     & 84.17                                                                     &                                                                                      & \cellcolor[HTML]{F0D999}C2S     & 80.39                                                                     \\ \cline{2-3} \cline{5-6} 
                                                                                     & \cellcolor[HTML]{DEE9A4}SIB     & 71.67                                                                     &                                                                                      & \cellcolor[HTML]{DEE9A4}SIB     & 82.76                                                                     \\ \cline{2-3} \cline{5-6} 
                                                                                     & \cellcolor[HTML]{DEE9A4}TextMix & \textbf{54.42}                                                                     &                                                                                      & \cellcolor[HTML]{DEE9A4}TextMix & 80.67                                                                     \\ \cline{2-3} \cline{5-6} 
                                                                                     & \cellcolor[HTML]{DEE9A4}SentMix & 57.09                                                            &                                                                                      & \cellcolor[HTML]{DEE9A4}SentMix & 81.09                                                                     \\ \cline{2-3} \cline{5-6} 
                                                                                     & \cellcolor[HTML]{DEE9A4}WordMix & 57.48                                                                     &                                                                                      & \cellcolor[HTML]{DEE9A4}WordMix & \textbf{76.91}                                                            \\ \cline{2-3} \cline{5-6} 
\multirow{-8}{*}{\textbf{DBpedia}}                                                   & \cellcolor[HTML]{72B6A1}INVSIB  & 77.79                                                                     & \multirow{-8}{*}{\textbf{\begin{tabular}[c]{@{}c@{}}Yelp\\ Polarity\end{tabular}}}   & \cellcolor[HTML]{72B6A1}INVSIB  & 84.32                                                                     \\ \hline
                                                                                     & \cellcolor[HTML]{DE7959}ORIG    & 75.64                                                                     &                                                                                      & \cellcolor[HTML]{DE7959}ORIG    & 99.25                                                                     \\ \cline{2-3} \cline{5-6} 
                                                                                     & \cellcolor[HTML]{F0D999}INV     & 69.71                                                                     &                                                                                      & \cellcolor[HTML]{F0D999}INV     & 90.01                                                                     \\ \cline{2-3} \cline{5-6} 
                                                                                     & \cellcolor[HTML]{F0D999}C2S     & 63.08                                                                     &                                                                                      & \cellcolor[HTML]{F0D999}C2S     & \textbf{65.15}                                                            \\ \cline{2-3} \cline{5-6} 
                                                                                     & \cellcolor[HTML]{DEE9A4}SIB     & 58.87                                                                     &                                                                                      & \cellcolor[HTML]{DEE9A4}SIB     & 84.48                                                                     \\ \cline{2-3} \cline{5-6} 
                                                                                     & \cellcolor[HTML]{DEE9A4}TextMix & \textbf{48.77}                                                            &                                                                                      & \cellcolor[HTML]{DEE9A4}TextMix & 78.42                                                                     \\ \cline{2-3} \cline{5-6} 
                                                                                     & \cellcolor[HTML]{DEE9A4}SentMix & 51.82                                                                     &                                                                                      & \cellcolor[HTML]{DEE9A4}SentMix & 79.45                                                                     \\ \cline{2-3} \cline{5-6} 
                                                                                     & \cellcolor[HTML]{DEE9A4}WordMix & 53.58                                                                     &                                                                                      & \cellcolor[HTML]{DEE9A4}WordMix & 72.64                                                                     \\ \cline{2-3} \cline{5-6} 
\multirow{-8}{*}{\textbf{\begin{tabular}[c]{@{}c@{}}Yahoo! \\ Answers\end{tabular}}} & \cellcolor[HTML]{72B6A1}INVSIB  & 62.17                                                                     & \multirow{-8}{*}{\textbf{IMDB}}                                                      & \cellcolor[HTML]{72B6A1}INVSIB  & 86.42                                                                     \\ \hline
\end{tabular}%
% }
\caption{Test suite accuracy (\%) by dataset and $TP$. Lower accuracy indicates higher defect detection potential. $TP$s are color coded by their variant type, where orange and light green are invariant and sibylvariant, respectively.}
\label{tab:testing_acc_full}
\end{table*}

\pagebreak

\begin{table*}[htbp]
\section{RQ3. Detailed Robustness Results}
\centering
\resizebox{\textwidth}{!}{%
\begin{tabular}{|c|
>{\columncolor[HTML]{DEE9A4}}c |c|c|c|c|
>{\columncolor[HTML]{DEE9A4}}c |c|c|c|}
\hline
\cellcolor[HTML]{EFEFEF}\textbf{Dataset}                                              & \cellcolor[HTML]{EFEFEF}\textbf{TP} & \cellcolor[HTML]{EFEFEF}\textbf{TF} & \cellcolor[HTML]{EFEFEF}\textbf{DWB} & \cellcolor[HTML]{EFEFEF}\textbf{TB} & \cellcolor[HTML]{EFEFEF}\textbf{Dataset}                                              & \cellcolor[HTML]{EFEFEF}\textbf{TP} & \cellcolor[HTML]{EFEFEF}\textbf{TF} & \cellcolor[HTML]{EFEFEF}\textbf{DWB} & \cellcolor[HTML]{EFEFEF}\textbf{TB} \\ \hline
                                                                                      & \cellcolor[HTML]{DE7959}ORIG        & 0.69                                & 0.56                                 & 0.54                                &                                                                                       & \cellcolor[HTML]{DE7959}ORIG        & \textbf{0.48}                       & 0.40                                 & 0.42                                \\ \cline{2-5} \cline{7-10} 
                                                                                      & \cellcolor[HTML]{F0D999}INV         & 0.73                                & 0.59                                 & 0.48                                &                                                                                       & \cellcolor[HTML]{F0D999}INV         & 0.49                                & 0.42                                 & \textbf{0.36}                       \\ \cline{2-5} \cline{7-10} 
                                                                                      & \cellcolor[HTML]{F0D999}C2S         & 0.66                                & 0.56                                 & 0.48                                &                                                                                       & \cellcolor[HTML]{F0D999}C2S         & 0.51                                & 0.49                                 & 0.50                                \\ \cline{2-5} \cline{7-10} 
                                                                                      & SIB                                 & 0.80                                & 0.69                                 & 0.49                                &                                                                                       & SIB                                 & 0.68                                & 0.55                                 & 0.63                                \\ \cline{2-5} \cline{7-10} 
                                                                                      & TextMix                             & 0.78                                & 0.60                                 & \textbf{0.45}                       &                                                                                       & TextMix                             & 0.56                                & 0.41                                 & 0.46                                \\ \cline{2-5} \cline{7-10} 
                                                                                      & SentMix                             & 0.70                                & 0.57                                 & 0.61                                &                                                                                       & SentMix                             & 0.58                                & 0.47                                 & 0.46                                \\ \cline{2-5} \cline{7-10} 
                                                                                      & WordMix                             & 0.84                                & 0.71                                 & 0.60                                &                                                                                       & WordMix                             & 0.74                                & 0.69                                 & 0.73                                \\ \cline{2-5} \cline{7-10} 
                                                                                      & $\alpha$TextMix                         & 0.77                                & 0.60                                 & 0.57                                &                                                                                       & $\alpha$TextMix                         & 0.55                                & \textbf{0.39}                        & 0.48                                \\ \cline{2-5} \cline{7-10} 
                                                                                      & $\alpha$SentMix                         & \textbf{0.60}                       & \textbf{0.43}                        & 0.46                                &                                                                                       & $\alpha$SentMix                         & 0.56                                & 0.49                                 & 0.53                                \\ \cline{2-5} \cline{7-10} 
                                                                                      & $\alpha$WordMix                         & 0.79                                & 0.64                                 & 0.55                                &                                                                                       & $\alpha$WordMix                         & 0.74                                & 0.69                                 & 0.69                                \\ \cline{2-5} \cline{7-10} 
\multirow{-11}{*}{\textbf{AG News}}                                                   & \cellcolor[HTML]{72B6A1}INVSIB      & 0.78                                & 0.62                                 & 0.57                                & \multirow{-11}{*}{\textbf{\begin{tabular}[c]{@{}c@{}}Amazon\\ Polarity\end{tabular}}} & \cellcolor[HTML]{72B6A1}INVSIB      & 0.65                                & 0.58                                 & 0.60                                \\ \hline
                                                                                      & \cellcolor[HTML]{DE7959}ORIG        & 0.92                                & 0.55                                 & 0.64                                &                                                                                       & \cellcolor[HTML]{DE7959}ORIG        & \textbf{0.48}                       & \textbf{0.20}                        & \textbf{0.28}                       \\ \cline{2-5} \cline{7-10} 
                                                                                      & \cellcolor[HTML]{F0D999}INV         & \textbf{0.76}                       & 0.47                                 & 0.48                                &                                                                                       & \cellcolor[HTML]{F0D999}INV         & 0.64                                & 0.41                                 & 0.52                                \\ \cline{2-5} \cline{7-10} 
                                                                                      & \cellcolor[HTML]{F0D999}C2S         & 0.85                                & 0.59                                 & 0.56                                &                                                                                       & \cellcolor[HTML]{F0D999}C2S         & 0.76                                & 0.58                                 & 0.66                                \\ \cline{2-5} \cline{7-10} 
                                                                                      & SIB                                 & 0.80                                & 0.58                                 & 0.64                                &                                                                                       & SIB                                 & 0.68                                & 0.53                                 & 0.65                                \\ \cline{2-5} \cline{7-10} 
                                                                                      & TextMix                             & 0.85                                & 0.48                                 & \textbf{0.41}                       &                                                                                       & TextMix                             & 0.76                                & 0.61                                 & 0.67                                \\ \cline{2-5} \cline{7-10} 
                                                                                      & SentMix                             & 0.96                                & 0.69                                 & 0.69                                &                                                                                       & SentMix                             & 0.70                                & 0.52                                 & 0.60                                \\ \cline{2-5} \cline{7-10} 
                                                                                      & WordMix                             & 0.91                                & 0.64                                 & 0.76                                &                                                                                       & WordMix                             & 0.78                                & 0.72                                 & 0.76                                \\ \cline{2-5} \cline{7-10} 
                                                                                      & $\alpha$TextMix                         & 0.82                                & 0.51                                 & 0.53                                &                                                                                       & $\alpha$TextMix                         & 0.61                                & 0.39                                 & 0.53                                \\ \cline{2-5} \cline{7-10} 
                                                                                      & $\alpha$SentMix                         & 0.87                                & \textbf{0.40}                        & 0.51                                &                                                                                       & $\alpha$SentMix                         & 0.94                                & 0.77                                 & 0.87                                \\ \cline{2-5} \cline{7-10} 
                                                                                      & $\alpha$WordMix                         & 0.83                                & 0.55                                 & 0.49                                &                                                                                       & $\alpha$WordMix                         & 0.62                                & 0.49                                 & 0.56                                \\ \cline{2-5} \cline{7-10} 
\multirow{-11}{*}{\textbf{DBpedia}}                                                   & \cellcolor[HTML]{72B6A1}INVSIB      & 0.83                                & 0.56                                 & 0.52                                & \multirow{-11}{*}{\textbf{\begin{tabular}[c]{@{}c@{}}Yelp\\ Polarity\end{tabular}}}   & \cellcolor[HTML]{72B6A1}INVSIB      & 0.75                                & 0.51                                 & 0.61                                \\ \hline
                                                                                      & \cellcolor[HTML]{DE7959}ORIG        & 0.54                                & 0.46                                 & 0.52                                &                                                                                       & \cellcolor[HTML]{DE7959}ORIG        & 0.86                                & \textbf{0.25}                        & 0.71                                \\ \cline{2-5} \cline{7-10} 
                                                                                      & \cellcolor[HTML]{F0D999}INV         & 0.57                                & 0.49                                 & 0.49                                &                                                                                       & \cellcolor[HTML]{F0D999}INV         & 0.70                                & 0.50                                 & 0.68                                \\ \cline{2-5} \cline{7-10} 
                                                                                      & \cellcolor[HTML]{F0D999}C2S         & 0.58                                & 0.53                                 & 0.54                                &                                                                                       & \cellcolor[HTML]{F0D999}C2S         & 0.93                                & 0.59                                 & 0.89                                \\ \cline{2-5} \cline{7-10} 
                                                                                      & SIB                                 & 0.56                                & 0.50                                 & 0.53                                &                                                                                       & SIB                                 & 0.71                                & 0.47                                 & 0.71                                \\ \cline{2-5} \cline{7-10} 
                                                                                      & TextMix                             & 0.58                                & 0.47                                 & 0.50                                &                                                                                       & TextMix                             & 0.85                                & 0.32                                 & 0.73                                \\ \cline{2-5} \cline{7-10} 
                                                                                      & SentMix                             & 0.72                                & 0.64                                 & 0.72                                &                                                                                       & SentMix                             & 0.80                                & 0.46                                 & 0.78                                \\ \cline{2-5} \cline{7-10} 
                                                                                      & WordMix                             & 0.65                                & 0.52                                 & 0.63                                &                                                                                       & WordMix                             & 0.84                                & 0.74                                 & 0.84                                \\ \cline{2-5} \cline{7-10} 
                                                                                      & $\alpha$TextMix                         & 0.54                                & 0.47                                 & 0.49                                &                                                                                       & $\alpha$TextMix                         & \textbf{0.56}                       & 0.32                                 & \textbf{0.55}                       \\ \cline{2-5} \cline{7-10} 
                                                                                      & $\alpha$SentMix                         & \textbf{0.48}                       & \textbf{0.41}                        & 0.48                                &                                                                                       & $\alpha$SentMix                         & 0.95                                & 0.91                                 & 0.96                                \\ \cline{2-5} \cline{7-10} 
                                                                                      & $\alpha$WordMix                         & 0.66                                & 0.59                                 & 0.61                                &                                                                                       & $\alpha$WordMix                         & 0.73                                & 0.52                                 & 0.68                                \\ \cline{2-5} \cline{7-10} 
\multirow{-11}{*}{\textbf{\begin{tabular}[c]{@{}c@{}}Yahoo! \\ Answers\end{tabular}}} & \cellcolor[HTML]{72B6A1}INVSIB      & 0.54                                & 0.44                                 & \textbf{0.46}                       & \multirow{-11}{*}{\textbf{IMDB}}                                                      & \cellcolor[HTML]{72B6A1}INVSIB      & 0.89                                & 0.79                                 & 0.88                                \\ \hline
\end{tabular}%
}
\caption{Attack success by dataset and $TP$ for three adversarial algorithms: TextFooler (TF), DeepWordBug (DWB), and TextBugger (TB). Lower attack success indicates higher adversarial robustness. $TP$s are color coded by their variant type, where orange and light green are invariant and sibylvariant, respectively.}
\label{tab:adv_robustness_full}
\end{table*}

\end{document}